%% file: main.tex
\documentclass{article}

\usepackage{PRIMEarxiv}

\usepackage[utf8]{inputenc} % allow utf-8 input
\usepackage[T1]{fontenc}    % use 8-bit T1 fonts
\usepackage{hyperref}       % hyperlinks
\usepackage{url}            % simple URL typesetting
\usepackage{booktabs}       % professional-quality tables
\usepackage{amsfonts}       % blackboard math symbols
\usepackage{nicefrac}       % compact symbols for 1/2, etc.
\usepackage{microtype}      % microtypography
\usepackage{lipsum}
\usepackage{fancyhdr}       % header
\usepackage{graphicx}       % graphics
\graphicspath{{media/}}     % organize your images and other figures under media/ folder

\usepackage{tabulary}
\usepackage{tabularx}
\usepackage{tabu}
\usepackage{array}
\usepackage{xcolor, colortbl} % To provide color for soul (for english editing), for adding cell color of table
\usepackage{pdflscape}
\usepackage{amsmath,amssymb}
\usepackage{ragged2e}
\usepackage{makecell}
\usepackage{rotating}
\usepackage{lscape}
\usepackage{pifont}
\usepackage{epstopdf}
\usepackage{colortbl}
\usepackage{silence}
\newcommand{\cmark}{\ding{51}}%
\newcommand{\xmark}{\ding{55}}%

\newcolumntype{P}[1]{>{\centering\arraybackslash}p{#1}}
\newcolumntype{M}[1]{>{\centering\arraybackslash}m{#1}}
\newcolumntype{L}{>{\centering\arraybackslash}m{3.9cm}}

\title{A Review of Radio Frequency Based Localization for Aerial and Ground Robots with 5G Future Perspectives
}

\author{
  Meisam Kabiri\footnotemark[1], ~Claudio Cimarelli\footnotemark[1], ~Hriday Bavle\footnotemark[1], ~Jose Luis Sanchez-Lopez\footnotemark[1],~ Holger Voos\footnotemark[1]\hspace{1.5mm}\footnotemark[2] \\
  \footnotemark[1] ~Interdisciplinary Center for Security Reliability and Trust (SnT), \\
  University of Luxembourg, L-1855 Luxembourg\\
  \footnotemark[2]~  Faculty of Science, Technology, and Medicine (FSTM), Department of Engineering, \\University of Luxembourg,
L-1359 Luxembourg\\
  \texttt{name.surname@uni.lu}
}

\begin{document}
\maketitle

\begin{abstract}
Efficient localization plays a vital role in many modern applications of Unmanned Ground Vehicles (UGV) and Unmanned aerial vehicles (UAVs), which would contribute to  improved control, safety, power economy, etc. The ubiquitous 5G NR (New Radio) cellular network will provide new opportunities for enhancing localization of UAVs and UGVs. In this paper, we review the radio frequency (RF) based approaches for localization. We review the RF features that can be utilized for localization and investigate the current methods suitable for  Unmanned vehicles under two general categories: range-based and fingerprinting. The existing state-of-the-art literature on RF-based localization for both UAVs and UGVs is examined, and the envisioned 5G NR for localization enhancement, and the future research direction are explored.
\end{abstract}

% keywords can be removed

%%%%%%%%%%%%%%%%%%%%%%%%%%%%%%%%%%%%%%%%%%

\input{Sections/introduction}

\input{Sections/section2}

\input{Sections/section3}

\input{Sections/section4}

\input{Sections/section5}

\input{Sections/section6}

\input{Sections/section7}

\input{Sections/section8}

\input{Sections/section9}

%%%%%%%%%%%%%%%%%%%%%%%%%%%%%%%%%%%%%%%%%%
\section{Conclusion}
 In this survey, we addressed RF-based localization mostly from a robotic point of view. First, we explored the methods that exist for RF-based localization extensively under two classes: Rage-based and Fingerprinting. Then we investigated and compared current state-of-the-art  RF-based localization applied to the robotic areas. Subsequently, the challenges and solutions that 5G will bring were discussed. Finally, the future research direction was given. 

5G NR will introduce features that could be harnessed to revolutionize localization and robots' applicability, like low latency, high throughput, and high-resolution angular and time-based measurements. However, 5G-based localization research is still in its infancy. Thus, many possibilities to exploit this novel communication technology in the robotic localization context are left unexplored. Nonetheless, the current solutions adopted for general RF receivers can be applied to 5G without modification. Up to now, cameras, IMU, LIDAR and GPS represent the predominant choice for building a SLAM system. However, the unique characteristics introduced by the 5G NR can establish RF-based localization among the most common robotic tools for safe autonomous navigation. Leveraging such un-exploited features and surpassing the main technological obstacles will be the focus of future research to make the integration of 5G into the current localization system seamless.

\section*{Acknowledgments}
This work was funded by the Fonds National de la Recherche of Luxembourg (FNR), under the projects C19/IS/13713801/5G-Sky. For the purpose of Open Access, the author has applied a CC BY public copyright license to any Author Accepted Manuscript version arising from this submission.

\bibliographystyle{unsrt}  
\bibliography{references}

\end{document}

%% file: Sections/introduction.tex
\section{Introduction}

Ground and aerial robots are essential for automating diverse civilian and commercial applications, such as search and rescue operations, connected and autonomous driving, precision agriculture, road traffic management, and many more potentials are yet to come~\cite{amponis2022drones, azmat2020potential}. For example, in the context of new communication technologies, drones may be effectively used as mobile 5G base stations for special events gathering crowds and solving network congestion. Nevertheless, in many of these scenarios, an autonomous robot navigates in a highly dynamic environment and requires precise self-location awareness to operate safely. Thus, localization can be regarded as the core of autonomous navigation systems. Common localization systems heavily depend on a Global Navigation Satellite System (GNSS), such as the Global Positioning System (GPS). While it provides sufficient accuracy in most situations, GPS-denied areas might impede or discourage relying solely on it. Due to the weakness of the GPS signal in indoor environments or forests, cameras, Inertial Measurement Units (IMU), and Light Detection and Ranging (LIDAR) are the principal alternatives to provide valuable information for enabling robot autonomous navigation, backed by mature theoretical studies and with many developed solutions. However, each of those sensors presents characteristics that make them unsuitable for specific environments or situations. Vision-based methods provide poor performance in low light, adverse meteorological conditions, or in presence of visual aliasing. IMU-based localization suffers from noisy measurements, and motion estimates drift rapidly. LIDAR provides the richest and most precise measurements of the surrounding 3D environment. However, it power consumption processing large-point cloud data. Moreover, in a wireless sensor network (WSN) with many nodes, GPS is not a reasonable solution regarding cost, power consumption, and form factor requirements of small Internet of Things (IoT) device \cite{bhat2020localization,hadir2017performance, chuang2014effective,kuriakose2014localization}. Therefore, multiple and diverse sensors must be integrated and fused to make the system robust.

As a complementary solution radio frequency (RF) signal can boost localization accuracy and robustness. Most RF metrics that have been used extensively include Time-of-Arrival (TOA), Time-Difference-of-Arrival (TDOA), Angle-of-arrival (AOA), and Received signal strength (RSS). Different technologies that can provide these measurement includes WiFi, Bluetooth, Ultra-Wideband (UWB), Zigbee, Radio Frequency Identification Device (RFID), Long Range Radio (LoRA), and cellular network \cite{zafari_survey_2019, maghdid_comprehensive_2021}. Cellular networks are beneficial because we can take advantage of the already deployed infrastructure covering many areas. 5G NR, comes with great potential for enhanced localization by not only providing accurate measurements and tools based on which even centimetre-level accuracy can be expected but also enabling the implementation of demanding fusion-based algorithms to compensate for the deficit of each sensor measurement through edge computing. The benefits of 5G NR for localization include but are not limited to: Wide area coverage, high angular and time resolution measurement for localization, increased probability of Line of Sight (LOS), and fast data rate. Concentrating on  UAVs and UGVs, 5G can revolutionize localization by providing:
\begin{itemize}
    \item Edge computing
    \item Vehicle to Everything communication (V2X)
    \item Beamforming
    \item Multi-array antenna
\end{itemize}

Several existing surveys exist in the area of localization covering the topic from different points of view, but also often overlap with each other \cite{yang2021survey, maghdid_comprehensive_2021, chowdhury_advances_2016, khelifi_survey_2019, tabassum_localization_2020, kumari_localization_2019, paul_localization_2017, dwivedi_positioning_2021, zafari_survey_2019, shakshuki_comparative_2019, perez_review_nodate}. For instance, Dwivedi et al.~\cite{dwivedi_positioning_2021} explore the positioning in 5G networks from the communication point of view. Yang et al. \cite{yang2021survey} review the positioning based on RF signal for UAVs with emphasis on the technologies and challenges, and Maghdid et al.~\cite{maghdid_comprehensive_2021} focus on the IoT technologies and performance metrics. Chowdhury et al.~\cite{chowdhury_advances_2016} provide a comprehensive survey for localization in WSN, which investigates the techniques and algorithms under diverse categories. Kumari et al. \cite{kumari_localization_2019} narrowed down the review to the localization in 3D space. Tabassum et  al. \cite{tabassum_localization_2020} present a brief overview of the localization approaches in WSN. Khelifi et al.~\cite{khelifi_survey_2019} review the localization for IoT under various categories of approaches. The reviews and the covered topics are compared in \autoref{tbl1}.
\begin{table}[h!t]
\caption{Comparison between existing reviews}
\label{tbl1}
% \begin{tabularx}{0.95\textwidth}{@{}1 *5{>{\centering\arraybackslash}X}@{}}
% \begin{tabularx}{0.8\textwidth}{[c]tbsss}
\begin{tabular}{  M{0.1 \textwidth}  M{0.6 \textwidth} M{0.02 \textwidth} M{0.02 \textwidth} M{0.02 \textwidth} M{0.02\textwidth}   } 

\Xhline{3\arrayrulewidth}
\bf Reference & \bf brief summary & \bf 1 & \bf 2 & \bf 3 & \bf 4 \\
\Xhline{3\arrayrulewidth}
% \rowcolor[gray]{0.9}
\cite{yang2021survey} &  A survey on RF localization for UAV with focus on technologies and performance metrics & \cmark & \xmark & \cmark& \xmark\\
\Xhline{1\arrayrulewidth}
\cite{maghdid_comprehensive_2021} &A survey on IoT localization, investigating technologies and performance metrics  & \xmark & \xmark & \cmark & \xmark \\
\Xhline{1\arrayrulewidth}
\cite{chowdhury_advances_2016} & A comprehensive survey on localization on WSN  & \xmark & \xmark & \cmark & \xmark \\
\Xhline{1\arrayrulewidth}
\cite{dwivedi_positioning_2021} & A overview of the 5G-based localization  & \xmark  & \xmark & \xmark & \cmark\\
\Xhline{1\arrayrulewidth}
\cite{khelifi_survey_2019} & A survey on IoT localization   & \xmark & \xmark & \cmark & \xmark \\
\Xhline{1\arrayrulewidth}
\cite{tabassum_localization_2020} & A survey on localization techniques on WSN  & \xmark & \xmark & \cmark & \xmark \\
\Xhline{1\arrayrulewidth}
\cite{kumari_localization_2019} & A survey on  localization in WSN in 3D space  & \xmark & \xmark & \cmark & \xmark \\
\Xhline{1\arrayrulewidth}
\cite{paul_localization_2017} & A survey on localization on WSN algorithm and techniques  & \xmark & \xmark & \cmark & \xmark\\
\Xhline{1\arrayrulewidth}
\cite{shakshuki_comparative_2019} & A brief  review on rang-free localization algorithms in WSN   & \xmark & \xmark & \cmark & \xmark\\
\Xhline{1\arrayrulewidth}
\cite{zafari_survey_2019} & A comprehensive survey of the indoor localization using different technologies    & \xmark & \xmark & \cmark & \xmark\\
\Xhline{1\arrayrulewidth}
\cite{perez_review_nodate} &  A brief review of technologies used for UAV positioning in indoor environment    & \cmark & \xmark & \xmark & \xmark\\
\Xhline{3\arrayrulewidth}
% \end{tabularx}\\

\end{tabular}\\
%  \vspace{2ex}
     {\hspace{3ex} \raggedright \fontsize{8}{12} \selectfont 1: UAV localization; 2: UGV localization; 3: Methods and algorithms; 4: 5G localization.   \\}
     
     \end{table}

Hence, this survey aims to address the issue that remained untouched by other state-of-the-art reviews. As mentioned, the existing studies focus on specific aspects of localization like technologies, methods, IoT applications, communication perspective, and classification of the approaches. Instead, this review covers the full extent of RF-based localization approaches, first by distinguishing the type of information that can be decoded from the received signal in the form of features and then by categorizing the methodologies based on the use,  range-based or fingerprinting (data-driven), of these RF features. Additionally, we address the use of RF for UAVs and UGVs applications and discuss 5G future potentials. Therefore, the main contributions of this review paper are as follows:

\begin{itemize}
\item Investigating the algorithms for RF-based localization that are highly likely to be used for challenging UGV and UAV applications.
\item Reviewing the existing works considered using RF specifically for  UAVs and UGVs positioning.
\item Discussing the new possibility that 5G NR will provide to cope with the current issues in  UAVs and UGVs localization problem.
\item Discussing the challenges for  ground and aerial robots localization and its integration with 5G NR. 
\end{itemize}

The survey paper is organized as follows: After introducing the mainly used RF features for localization, we briefly introduce localization techniques and how they can be used. Then we proceed to explore the up-to-date, state-of-the-art algorithms that have been widely used in RF-based localization. In the next section, we concentrate on the available RF-based localization applied to  UAVs and UGVs. The 5G NR potentials and benefits expected to be brought to the  aerial and ground vehicles localization are also explained. We also review the current state-of-the-art localization based on 5G NR, and finally, the challenges and future direction will be explored.

%% file: Sections/section2.tex
\section{RF Features}
The received radio frequency signal involves features that encode information about the transmitter's and receiver's relative position. Therefore, in this section, we describe the most relevant and widely used for localization purposes.

\subsection{Received Signal Strength} Received Signal Strength (RSS) is one of the most used features for localization simply due to its low hardware requirements and easy implementation compared to other counterparts. RSS is the received power at the receiver over the bandwidth. The main concept behind using RSS for localization is that the power attenuation of the signal from transmitter to receiver depends on the distance. Extraction of the exact relation, however, in real environments seems infeasible due to the unknown channel model, and the majority of existing literature relies on using simple models for mapping RSS to distance or range. The most popular model is the log-distance one-slope propagation model:
    \begin{align}\label{path_loss}
    P_{d}[dB] =&  P_0[dB]-10\alpha log_{10}(d/d_0)%\\\nonumber
    +X_\sigma[dB]+b[dB]\,, \\\nonumber
    d =& ||X - S||\,,
\end{align}
where:
\begin{itemize}
    \item $P_0 $:  power at the reference distance $d_0$ from the transmitter (usually 1m)
    \item $P_{d}$:  received power at distance $d$ from the transmitter
    \item $X_\sigma$:  shadowing effect( mostly considered as Gaussian)
    \item  $\alpha$: Path loss exponent (PLE), the rate at which power decrease over distance
    \item $b$:  bias error
\end{itemize}
A more complex form of propagation model has been used, like two-slope, third-order, and higher up to the sixth-order model~\cite{tian_third-order_2013,martinez-sala_accurate_2005}. Lee et al.~\cite{lee_genetic_2021} propose to find the best model for each transmitter-receiver pair using the Genetic algorithm to search between multi-state path loss models with k states. It should be noted the transmission power is assumed to be fixed. When power control is applied at the transmitter, which might be the case in 5G, range inference based on distance is impossible. Transmission power and PLE are usually obtained by doing a pre-test, collecting data from the environment and matching the suitable values based on the model. Algorithms that deal with unknown PLE and transmission power thus offer the advantage of removing the need for an intensive pre-test phase.

Another contributing reason for less accurate RSS-based localization -in addition to channel modeling error- is that RSS is not stable and fixed over time and under changing environments. In \cite{lin_microscopic_nodate}, three main factors that play an important part in unstable RSS readings are discussed, and the effect of antenna orientation, foreground obstacles, and beacon density are explored.

\subsection{Time-Of-Arrival} 
Time-Of-Arrival (TOA) technique measures the time it takes for an RF signal to travel from transmitter to receiver, multiplying by the signal's speed in the medium, which is usually the propagation speed of light, and range can be inferred. As the obstruction does not impact the speed of the wave that much,  TOA delivers higher accuracy compared to RSS provided that tight clock synchronization among receiver and transmitters is done and there is Line-of-sight (LOS) path. The existence of LOS and clock synchronization, however, are the two challenges in TOA estimation. When no LOS path exists, the multi-path components are received while each travels  further than the distance between the receiver and transmitter. Stringent clock synchronization also calls for complex hardware.

\subsection{Time-Difference-Of-Arrival} Time-Difference-Of-Arrival (TDOA) is another time-based measurement which is the difference between the time signal travelled from two transmitters to the same receiver. In the conventional approach,  cross-correlation is used to extract this value, computing the delay that maximizes the cross-correlation function. One advantage of TDOA compared to TOA is that synchronization is only required among the transmitters. TDOA suffers from the same issues as TOA, i.e. imperfection hardware and LOS blockage.

\subsection{Angle-Of-Arrival}  Angle-Of-Arrival (AOA) method for localization has not been incorporated as much as other counterparts. The use of directional antennas, multi-element arrays for MIMO, and mmWave especially for 5G cellular networks has attracted more attention. Similar to TOA, AOA also suffers from LOS blockage. Other multi-path components from NLOS, can lead to erroneous and misleading information about the direction of the arrival. Thus, AOA is usually fused with other data for localization purposes.

\subsection{Channel System Information}
Compared to the mentioned RF features, Channel System Information (CSI) provide much richer information, such that all the other values can be extracted by analysing it. CSI gives information about the communication channel, e.g., fading, scattering, and power decay, and how the propagating signal is impacted at a specific carrier frequency at each path. CSI amplitude and phase are dependent on the poses of  the transmitter, receiver, and the surrounding environment with objects therein.  Compared to RSS, TOA, TDOA, and RSS which give one value per measurement, CSI consists of a matrix with each entry representing the Channel Frequency Response (CFR). Thus, there is much more data to deal with and analyse. Moreover, it is raw information that is not that intuitive compared to the other measurement. Hence machine learning algorithms and fingerprinting are suitable candidates for CSI applications.

%% file: Sections/section3.tex
\section{Overview of RF-based Localization Techniques}

Loosely speaking, localization based on radio signals can be divided into range-based and fingerprinting methods.

\begin{itemize}
    \item {\bf Range-based techniques:} Localization is done by inferring the distance or angle of the target from some node based on the measurements. Time-Of-Arrival(TOA), Time-Difference-Of-Arrival (TDOA), and Received Signal Strength (RSS) provide ranges, while Angle-Of-Arrival (AOA) gives bearings measurements. In a sensor framework, two or several of these methods can be combined, which might result in a better outcome. In the next stage, the extracted ranges or bearings are used to estimate the location taking advantage of various mathematical tools like Maximum Likelihood (ML), Least Squares (LS) approach, Bayesian model, or different types of filters like Kalman filter(KF), extended Kalman (EKF), Unscented Kalman filter (UKF), and Particle filter (PF). In the next section, we will explain the methods used for range-based localization. 
    
    \item {\bf Range-free or Fingerprinting:} Instead of calculating the distance or direction, the environmental survey is performed to obtain fingerprints or features recorded on a database like the location-RSS pair's value, and then in on-online mode, for every new measurement, the localization is performed by finding the best match in the data-set. More generally, this method consists of mapping and matching. Compared to the range-based approach, fingerprinting techniques are more accurate and demanding to implement, requiring a pretest to create an extensive database. In addition, fingerprinting methods differ in generating and updating the data set and the matching process. Nevertheless, fingerprinting is widely used for CSI and RSS-based localization. 
\end{itemize}

\begin{figure}
\centering
\includegraphics[width = 0.4 \textwidth]{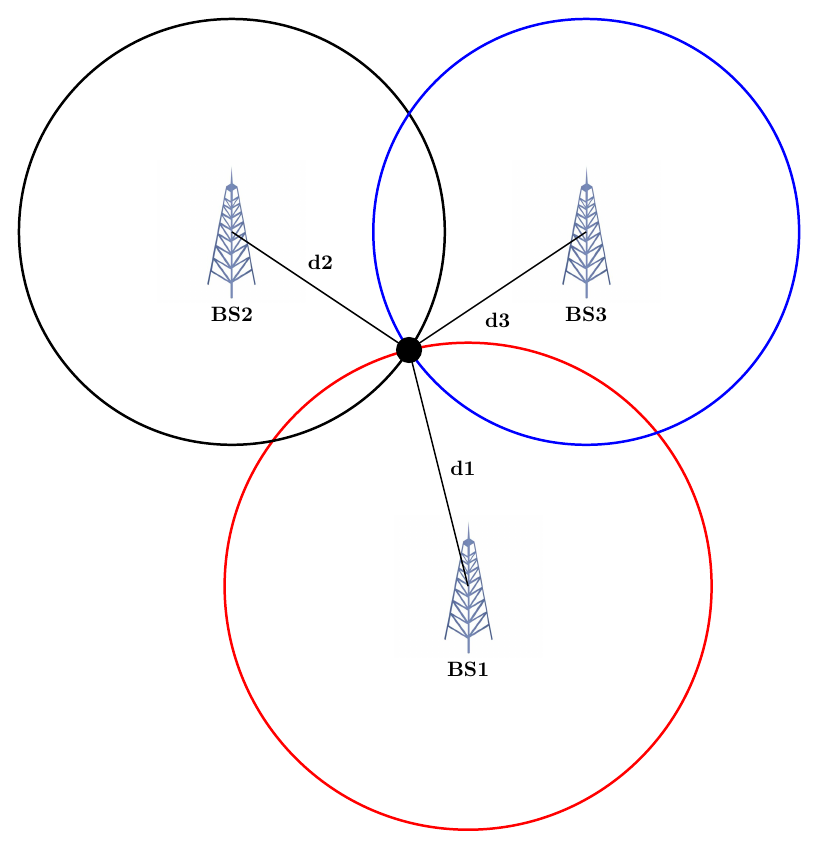}
\caption{Trilateration: localization based on the range from 3 anchors (TOA, RSS).}
\label{lateration}
\end{figure}

\begin{figure}
\centering
\includegraphics[width = 0.4 \textwidth]{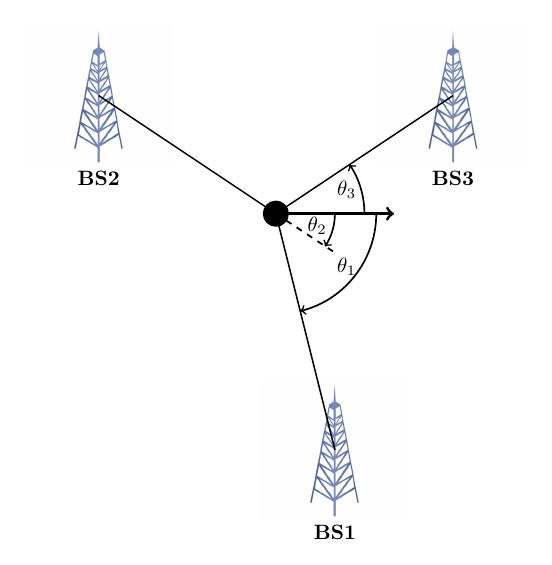}
\caption{Triangulation: localization based on AOA from 3 BS.}
\label{triangulation}
\end{figure}

\begin{figure}
\centering
\includegraphics[width = 0.4 \textwidth]{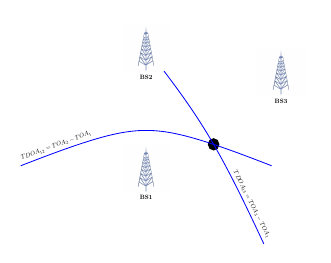}
\caption{Localization based on TDOA, the intersection of the hyperbolas.}
\label{tdoa}
\end{figure}

%% file: Sections/section4.tex
\section{Range-based Algorithms}
%%%%%%%%%%%%%%%%%%%%%%%%%%%%%%%%%%%%%
Having extracted ranges or bearing from the RF signal, intuitive mathematical or geometrical approaches are then leveraged to infer the receiver's position. The most important methods include Multi-Lateration/Triangulation, Min-Max, Multidimensional scaling (MDS), Least Squares (LS), Maximum Likelihood (ML), Bayesian inference method, and Bayesian Filters.  \autoref{tab:range-based-comp} gives a comparison of range-based methods.

\subsection{Multi-Lateration/Triangulation}
Based on the ranges, angles, or range differences a set of equations can be constructed which usually results in an over-determined set of non-linear equations. For N number of BS position at $S_i  = [S_{ix} , S_{i_y}]^T$ and the target at location $X = [X_x X_y]^T$, we can write

\begin{align}
d_i &= ||X - S_i|| +\sigma, \quad i = 1,2,.., N\,,\\
\delta d_i  &= ||X - S_i - S_1||+\sigma, \quad i = 2,..., N \,,\\
a_i & = \arctan (\frac{X_y - S_{iy}}{X_x - S_{ix}})+\sigma, \quad 1,..., N \,, 
\end{align}

where $\sigma$ is the measurement error, $d_i$, $\delta d_i$, and $a_i$ are the range, range-difference, and angle measurements. Geometrically speaking, in the simplest scenario, having the range from target to three Base stations (BS) in 2D, the intersection of 3 circles with the centre on the BS and radius of the measured range will be the solution (see \autoref{lateration}). In 3D, the intersection of 4 Spheres is the response. For TDOA, the hyperbolas can be constructed with the focus being the two BS (see \autoref{tdoa}). In the AOA framework, the measured angles with the help of the geometric properties of the triangle could determine the target location (see \autoref{triangulation}). In an ideal case, these methods intersect in one single point. However, this never happens in the real world, which gives rise to several scenarios \cite{le_geometric_2021}.

The set of equations can then be solved either in approximated closed-form or iterative way. Closed-form formulation leads to an easy, low-complexity solution. For example, in \cite{norrdine_algebraic_2012}, a closed-form algebraic solution for a target localization for both trilateration (three reference points) and multi-lateration (more than three reference points) is solved. The uncertainty of each information piece is considered by adding a variance matrix to the equations. However, it assumes the known covariance matrix error which is not realistic. In an extended version of this method~\cite{booranawong_rssi-based_2021}, after applying the standard multi-lateration procedure, if the found solution lies within the reference node positions, it is considered the final solution. Otherwise, the algorithm keeps searching for the solution inside the zones that are determined within the reference nodes based on the strength of the RSS level (the zone closer to the base station from which the strongest RSS level is read has the highest probability of containing the solution). In the search process for a solution inside a zone, virtual positions are defined inside zones, and then the final target location is selected based on the distance between the first estimation and the virtual positions.

\subsection{Min-Max}

\begin{figure}
\centering
\includegraphics[width = 0.3 \textwidth]{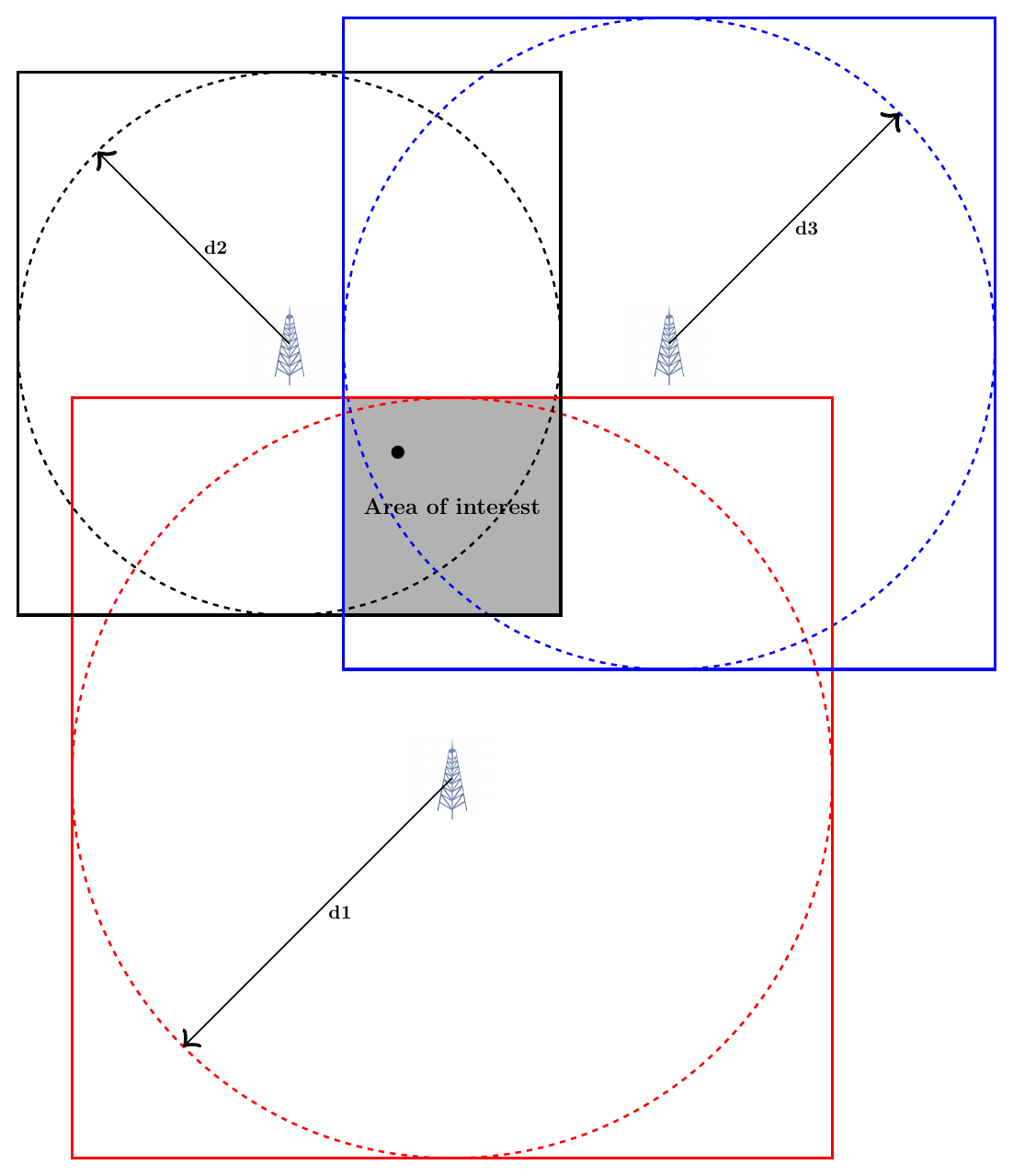}
\caption{Min-Max algorithm.}
\label{min-max}
\end{figure}

Min-Max is a simple, intuitive, and geometrical-based technique when an easy implementation is desired \cite{janssen_comparing_2020,yang_rss-based_2021,monta_evaluation_2016}. According to ranges, squares are formed that circumscribe the circles around each BS with radius $d_i$ as the distance between the BS and target. Then the vertices of a rectangle called the area of interest are found as shown in \autoref{min-max}. In the simplest version, the centroid of this rectangle is selected as the estimated position. Among Min-Max variants, Extended Min-Max \cite{xie_improved_2014} uses the weighted centroid instead of the geometric centroid. Furthermore, Yang et al. \cite{yang_rss-based_2021} introduce a new strategy with partition area.

\subsection{Multidimensional scaling (MDS)}
Multidimensional scaling (MDS) is a computationally efficient approach, especially for a cooperative framework consisting of a group of nodes for localization that due to high dimensional search space, finding the optimized value is demanding. MDS is a visualization technique by which the pairwise range could be mapped to the lower dimensional cartesian space that can be graphically displayed. MDS method offers an analytical closed-form solution which makes it advantageous in terms of computational burden, efficiency, and eases of implementation \cite{seco_survey_2009, saeed_state---art_2019}.

\subsection{Least squares (LS)}
Least squares (LS) is a well-known method to estimate an unknown. It can be formulated as finding a solution that minimizes the square of the error such that:

\begin{align}
&\underset{X}{\arg\min} \sum_{i =1}^{N}  \left( RSS_i - \left(P^i_0-10\alpha log_{10}(||X-S_i||)\right)\right)^2 \,,\\
&\underset{X}{\arg\min} \sum_{i =1}^{N}  \left( TOA_i c  -   ||X-S_i||\right)^2 \,,\\
&\underset{X}{\arg\min} \sum_{i =2}^{N}  \left( TDOA_i c -   ||X-S_i||+||X-S_1||\right)^2 \,,\\
&\underset{X}{\arg\min} \sum_{i =1}^{N}  \left( AOA_i  -  \arctan (\frac{X_y - S_{yi}}{X_x - S_{xi}})\right)^2\,,
\end{align}
where c is the speed of the light, $RSS_i$, $TOA_i$,  $TDOA_i$, and $AOA_i$ are the measurements with respect to the ith BS. For the cooperative case where there are M targets to be localized, it can be written as :

\begin{align}
&\underset{X}{\arg\min} \sum_{j =1}^{N}\sum_{i = 1}^M  \left( RSS_{ij} - \left(P_{0i}-10\alpha log_{10}(||X_j-S_i||)\right)\right)^2 \,,\\
&\underset{X}{\arg\min} \sum_{j =1}^{N}\sum_{i = 1}^M  \left( TOA_{ij}c -   ||X_j-S_i||\right)^2\,,\\
&\underset{X}{\arg\min} \sum_{j =1}^{N}\sum_{i = 1}^M  \left( TDOA_{ij} c-   ||X_j-S_i||+ ||X_j-S_1||\right)^2\,,\\
&\underset{X}{\arg\min} \sum_{j=1}^{N}\sum_{i =1}^{M}  \left( AOA_{ij}  -  \arctan (\frac{X_{yi} - S_{yi}}{X_{xi} - S_{xi}})\right)^2\,.
\end{align}

A straightforward solution to cope with the highly non-linearity is linearization, like using Taylor expansion \cite{zhai_recursive_2021}. However, the recursive LS method brings more fidelity and accuracy in the cost of complexity and computation burden \cite{zhai_recursive_2021}. 
The Weighted Least Squares (WLS) method is more efficient, which puts different weights on the measurement. In one intuitive way, weights are selected based on the distance. The optimal way is to exploit the covariance of the measurement noise for WLS. When such information is not at hand for noise, there are some efforts to estimate it \cite{kang_hybrid_2020}. 

In addition to position, other unknown parameters can also be jointly estimated, like unknown PLE \cite{li_rss-based_2006} and transmission power (for RSS-based localization) \cite{sun_robust_2021}. Recent work \cite{sun_robust_2021} has expanded this concept further by considering both unknown weight matrices and Non-Line-of-Sight (NLOS) impact, modeled by introducing random variables to the propagation model. While this approach enhances accuracy, it significantly increases problem complexity. To address this challenge, the semi-definite relaxation (SDR) technique serves as an effective approximation, transforming the non-convex problem into convex semi-definite programming.

In a more efficient framework, the relative error is employed instead of the estimation error, called Least Squares Relative Error (LSRE) \cite{wang_cooperative_2019, shi_least_2020}. The relative error is the ratio of the absolute error to the measured value. In LS, all observations are treated equally, implying the same precision for all data, which is critical in reality.

\subsection{Maximum Likelihood (ML)}
Maximum Likelihood is one of the most used approaches for localization, resulting in a non-convex non-linear optimization problem. The object of ML is to maximize the likelihood function:
\begin{gather}
\underset{X}{\arg\max} \quad  p (X|z) \,.
\end{gather}

The LS minimization of the observation errors normalized by measurement variances gives the ML solution. 

\begin{align}
&\underset{X}{\arg\min} \sum_{j =1}^{N}\sum_{i = 1}^M  \left(\frac{ RSS_{ij} - \left(P_{0i}-10\alpha log_{10}(||X_j-S_i||)\right)}{\sigma_{ij}}\right)^2\,,\\
&\underset{X}{\arg\min} \sum_{j =1}^{N}\sum_{i = 1}^M  \left(\frac{ TOA_{ij}c -   ||X_j-S_i||}{\sigma_{ij}}\right)^2\,,\\
&\underset{X}{\arg\min} \sum_{j =1}^{N}\sum_{i = 1}^M  \left( \frac{TDOA_{ij} c-   ||X_j-S_i||+ ||X_j-S_1||}{\sigma_{ij}}\right)^2 \,,\\
&\underset{X}{\arg\min} \sum_{j=1}^{N}\sum_{i =1}^{M}  \left( \frac{AOA_{ij}  -  \arctan (\frac{X_{yi} - S_{yi}}{X_{xi} - S_{xi}})}{\sigma_{ij}}\right)^2\,.
\end{align}
Solving it is not trivial and imposes a high computational cost. Subsequently, variant relaxations criteria are used to deal with the complexity like semi-definite programming (SDP) relaxation and second-order cone programming (SOCP) relaxation to approximate the complex problem, or numerical approaches are employed like Newton-Raphson. In a joint-ML scheme, location along with channel parameters are estimated \cite{zemek_joint_2007,coluccia_ml_2010}. \cite{coluccia_ml_2010} also explores the combination of multi-lateration with ML to make a compromise between performance and complexity where multi-lateration is exploited for location estimation, and ML is utilized for channel parameters estimation. The paper then compares it with the joint ML method. The effect of the fault and various noises are taken into account in \cite{mei_rss-based_2021, jiang_expectation_2021}. \cite{mei_rss-based_2021} addresses byzantine fault and NLOS effect. Byzantine fault is considered by including non-Gaussian interference noise corrupting the transmission data, and the NLOS effect is modelled by adding a bias term to the propagation model. First-order Taylor series expansion is utilized to simplify the model so that some transformation turns it into a generalized trust-region sub-problem (GTRS). Additive and multiplicative noises are handled in \cite{jiang_expectation_2021} using ML, taking advantage of iterative expectation-maximization.

\subsection{Bayesian Inference Method}
 The Bayesian approach gives a distribution, instead of an estimated value, as an outcome which is more informative than LS. Compared to ML, where parameters are considered fixed, in Bayesian method they are treated as random variable with known prior distribution \cite{phoong2015comparison}. Based on Bayes's theorem, the position and the parameters act as random variables where the prior information and observations are leveraged to infer and update the posterior distribution of unknown random variables. For example, in \cite{jin_bayesian_2020}, PLE and position are treated as mutually independent random variables from which the posterior distributions are derived. To do so, a message passing algorithm, called belief propagation \cite{pearl1988probabilistic}, is used on the factor graph, which allows for efficient computation of marginal distributions and dealing with the problem's intractability. The cooperative localization scenario is also dealt with using the Bayesian method in \cite{jin_bayesian_2020}. In this paper, in addition to PLE, transmission power is estimated.

\subsection{Bayesian Filters}
When estimating the dynamic states, which is the case in robotic applications, filters can work effectively. They work based on alternating between two steps, i.e., prediction and update. For example, the classic Kalman filter is the optimal estimator for a linear system in the presence of Gaussian noise. They can also be employed readily to fuse the observations of various sensors. In robotics, the movement dynamic is considered constant velocity or constant acceleration, which usually performs acceptably in practice. 

While the process model is linear, the observation model becomes non-linear. In this case, sup-optimal non-linear filters are exploited like Extended Kalman filter \cite{benini_imuuwbvision-based_2013}, Particle Filter \cite{wu_particle-filter-based_2014}, and Unscented Kalman Filter~\cite{yin_uwb-based_2016}. The Extended Kalman filter is the first-order Taylor series expansion of the non-linearity widely used in diverse applications with an acceptable outcome. However, such a first-order approximation might introduce errors in the posterior distribution estimation \cite{wan_unscented_2000}. The unscented Kalman filter (UKF) and particle filter (PF) can outperform EKF by removing the linearization step and carrying out deterministic sampling. Based on unscented transform (UT), UKF transforms states to weighted sigma points based on which prediction and update are executed. PF performs sequential importance sampling, drawing the particles and their corresponding weights from the probability density. We will review works based on Bayesian filters in \autoref{rob_app}.

\begin{table}[h]
\caption {Comparison between range-based methods.}
\label{tab:range-based-comp}
    \begin{tabular}{  L L L L }
    
        \toprule
\textbf{Range-based methods}   & \textbf{Scenario}  & \textbf{Advantages} & \textbf{Disadvantages}
\\\midrule
\rowcolor{gray!15} Multi-lateration/ Triangulation 
& fast and rough estimation scenarios       &  simple calculation
 & limited accuracy,  sensitive to measurement error
\\

 Min-Max  &  fast and rough estimation scenarios   &   low complexity, easy implementation  & limited accuracy \\

 \rowcolor{gray!15} Multidimensional Scaling (MDS)          &  cooperative localization         & reduce the complexity  & difficult to include the knowledge about unequal measurement error\\
% \hline

Least Square (LS)   & high accuracy   & easier implementation and less demanding than ML and Bayesian,  gives estimation uncertainty &   computationally demanding,   less optimal compared to ML and Bayesian \\
% \hline

 \rowcolor{gray!15} Maximum Likelihood (ML)   & high accuracy,   inaccurate  prior information (outperform Bayesian) &    gives estimation uncertainty    &  computationally demanding  \\
 % \hline

 Bayesian Inference   & higher accuracy,   sparse observations &  gives estimation uncertainty   &  computationally demanding (more demanding than LS and ML)  \\
 % \hline

 \rowcolor{gray!15} Extended Kalman Filter (EKF)  & real-time dynamic state estimation   Easy implementation for real-time                & simpler multi-sensor fusion,  suitable for mobile targets,  easy implementation,  gives estimation uncertainty & not useful for non-gaussian noise,  less optimal compared to UKF and PF       \\
 % \hline
 
 Unscented Kalman filter (UKF)   & real-time dynamic state estimation,   better accuracy compared to EKF             & simpler multi-sensor fusion,  suitable for mobile target,  gives estimation uncertainty & not useful for non-gaussian noise      \\
 % \hline

\rowcolor{gray!15} Particle Filter (PF)  & high accuracy dynamic state estimation     &  handling non-gaussian noise,  gives estimation uncertainty & computationally demanding,  difficult implementation \\

        \bottomrule
    \end{tabular}
\end{table}

%% file: Sections/section5.tex
\section{Fingerprinting}

As an alternative to the range-based approach, fingerprint methods are based on the collected data to infer the position instead of relying on the model. This method can better handle the error caused by the modelling errors and noises. Fingerprinting is mostly used for RSS and CSI-based localization, but it has also been exploited for other RF-based measurement \cite{schmitz_tdoa_2015, tan_uav_2020, yu_fingerprinting_2011, ha_lora_2019, de_sousa_applying_2019} and also for a hybrid scheme \cite{wei_joint_2020, he_hybrid_2020, li_crlb-based_2019}. 
%For example, in a TDOA-based framework, \cite{de_sousa_applying_2019} employs estimated time delay along with other combinations of the multi-path components.  \cite{chen_aoa-aware_2020} recorded AOA derived from Channel State Information (CSI) phase, considering all possible combinations of the transmitter antennas, receiver antennas, and OFDM sub-carriers. 

\begin{figure}
\centering
\vspace{1mm}
\includegraphics[width = 0.9 \textwidth]{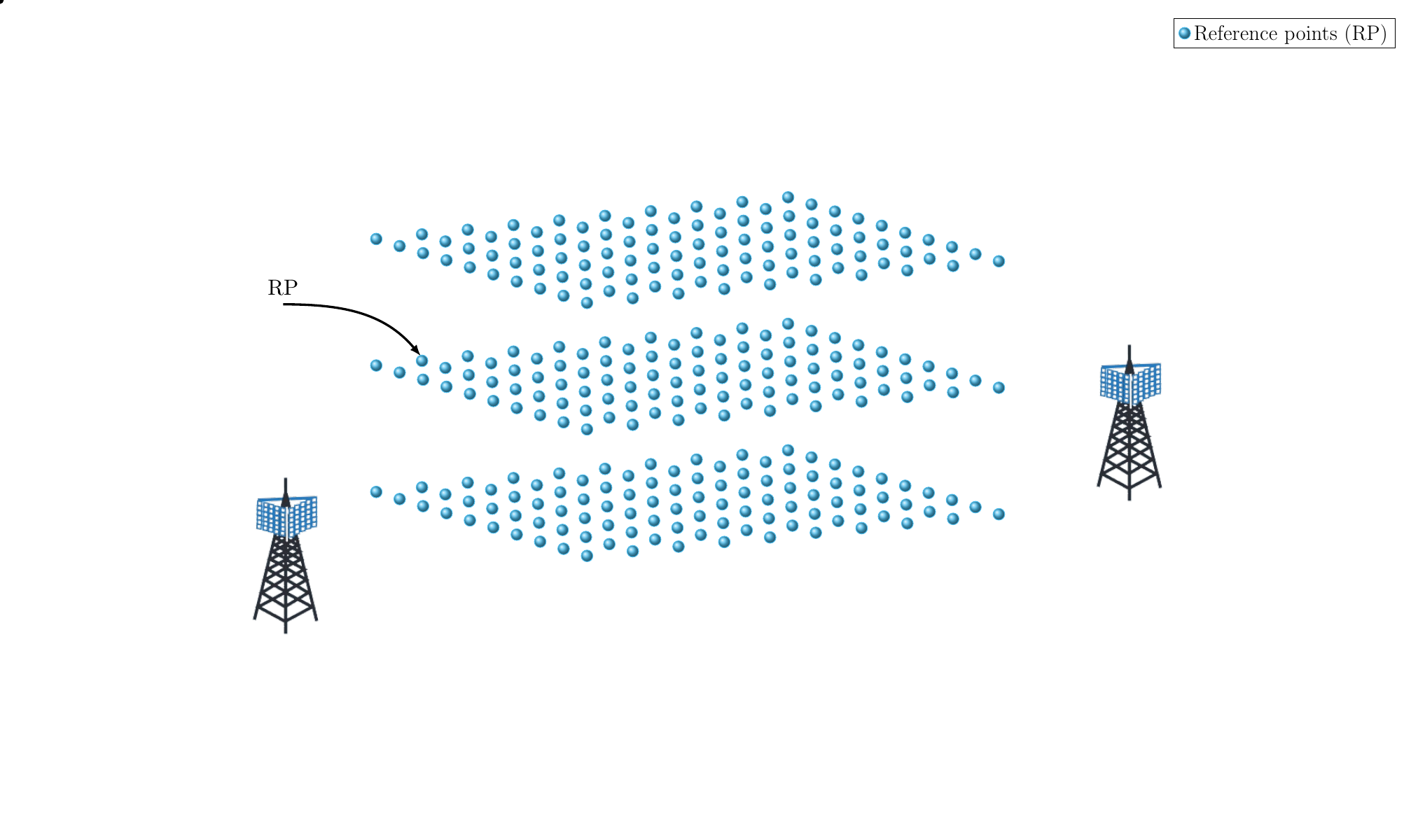}
\caption{Reference points in 3D.}
\label{rp}
\end{figure}

\begin{figure}
\centering
\includegraphics[width = 0.9 \textwidth]{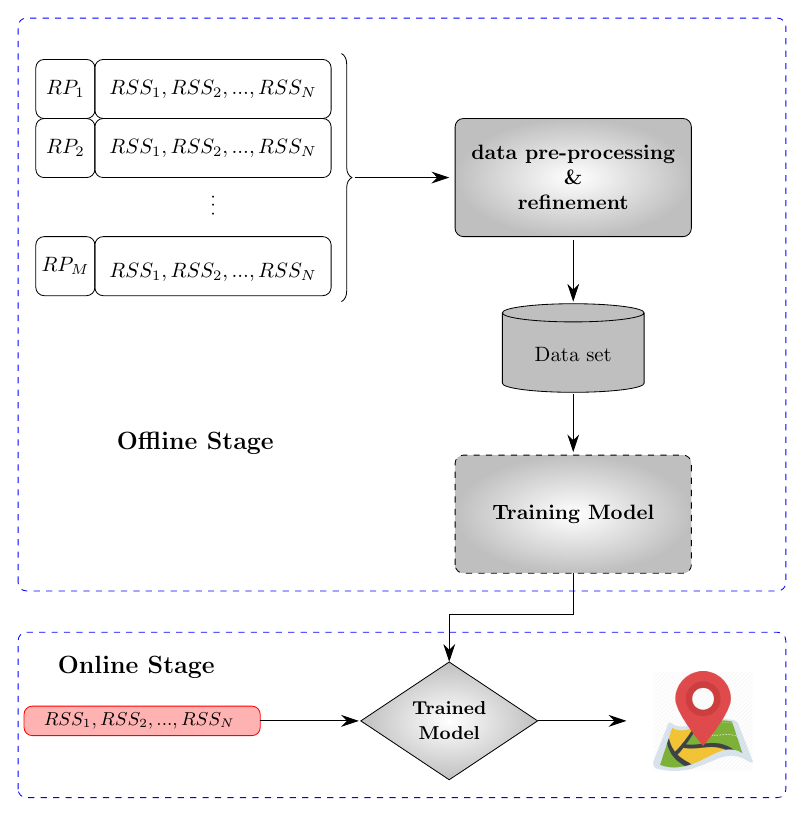}
\caption{Fingerprinting block diagram.}
\label{diag}
\end{figure}

This method usually results in more accurate estimation and comes at the cost of the laborious step of data collection. Generally speaking, fingerprinting consists of an offline mode and an online mode. In the offline step, a data set consists of recorded measurements or RF features, and ground truth position at references points (RP) (see \autoref{rp}), which are called fingerprints is generated based on which, in the online phase, estimation is made through matching (see \autoref{diag}). 

\subsection{Offline step}

One of the main challenges in the fingerprinting approach is collecting and maintaining a proper fingerprint database. The most straightforward approach to generating the radio map or data set is to measure and record the fingerprints over all areas. However, due to the ample space, it is not practical to do detailed measurements and costly site surveys. Moreover, the fingerprint map becomes outdated due to the radio signal features and dynamic environment. This problem is discussed in \cite{talvitie_distance-based_2015}, where the effect of incomplete data set is explored by applying several interpolations and extrapolations to recover the missing data. 
%, which is simulated by artificially removing some data from a data set gathered. 
The work of \cite{bi_fast_2019} explores deterministic adaptive path loss model interpolation for radio map generation focusing on improving the fastness of the process and decreasing the time of radio map construction. Each AP's RSS path loss model is extracted using available reference points, LS, and interpolation. Gaussian regression is another probabilistic method for this purpose. \cite{sun_augmentation_2018} exploits Gaussian Process Regression (GPR) to predict the RSS spatial distribution based on the available data set. To do so, instead of using a basic zero mean function and a single squared exponential as kernel function for GPR, compound GPR kernel functions are used. In \cite{yiu_wireless_2017}, more information regarding radio map generation techniques can be found. Experiments in the real world also make the comparison.

It is worth mentioning that most fingerprinting localization exploits the collected raw data or their scaled value. However, this is not efficient in the presence of heterogeneity. The heterogeneity of device pairs, i.e., transmitter and receiver, can affect the RF-based localization, as each device use different hardware parameter, like antenna gain for transmitters. The different signatures might be recorded in the same situation for each transmitter-receiver combination. As a remedy to this issue, other variants of perceived measurements are used. For example, differences in signal strengths are used in~\cite{tiwary_differential_2021} to address both device heterogeneity and temporal variation of RSS. Another differential RSS-based localization is handled in \cite{sun_improved_2020}. Finally, authors in \cite{kjaergaard_indoor_2011} present hyperbolic location fingerprinting in which fingerprints are recorded as RSS log ratios between pairs of base stations instead of absolute RSS.

\subsection{Online phase}
Machine learning algorithms are mainly used for matching and estimation, including K Nearest Neighbor (KNN), Weighted K Nearest Neighbor (WKNN), Artificial neural network (ANN), Convolutional neural network (CNN), Support Vector Machine (SVM), and Random Forest.

\subsubsection{Classical Machine Learning}

% \subsubsection{K Nearest Neighbor (KNN)}
K-Nearest Neighbor (KNN) is the first and simple candidate to select K points in the data set based on similarity. Considering the computation complexity, the achieved performance for this algorithm is acceptable. In a WKNN, different weights are assigned to measurements based on criteria such as the distance to the AP. For example, in \cite{fang_optimal_2018}, WKNN is employed where based on the inverse of RSS distance, weights are assigned to the APs. This criterion, however, is not consistent with position distance due to the nature of path loss which decays logarithmically based on distance (the reader is referred to \cite{wang_novel_2020} for more details). To address this issue, \cite{wang_novel_2020} introduces a new weighting process and a new distance measure based on the RSS similarity and spatial position.

One disadvantage of traditional KNN is that it is implemented with a fixed number of K. A Weighted Adaptive KNN Algorithm is adopted in \cite{zhang_weighted_2022} to settle this problem which can choose a variable number of RPs according to both the improved RSS similarity and position proximity. Density-based spatial clustering of applications with noise (DBSCAN) method and Affinity Propagation Clustering (APC) also does not require the number of clusters to be pre-determined and is suitable for handling large databases. The authors in \cite{liu_clustering-based_2021} show that clustering could reduce noise's impact in large data sets. Hence, the DBSCAN method is proposed for localization, where the fingerprints are divided around a centre point based on the density. Li et al. \cite{li_integrated_2016} apply APC offline until the algorithm converges to a final clustering after several iterations and message passing among the points (attraction and attribution messages). 

Random Forest is a classifier and regression method based on multiple decision trees that enjoy fast training and prediction, which works well with high dimensional data. Thus it is a suitable candidate for handling large data-set \cite{de_sousa_applying_2019, pandey_sele_2021}. SVM is another classifier engine that can resolve the regression problem. For localization, examples of multi-class SVM can be seen in \cite{wang_target_2021, chriki_svm-based_2017}. To make the fingerprints more separable, \cite{wang_target_2021} exploits a fuzzy kernel that maps the data to the higher dimension space. 

% \subsection{Support Vector Machine (SVM)}
% SVM is a classifier engine that can resolve the regression problem. For localization, examples of  multi-class SVM  can be seen in \cite{wang_target_2021, chriki_svm-based_2017}. %To make the fingerprints more separable, \cite{wang_target_2021} exploits a fuzzy kernel that maps the data to the higher dimension space. 

% \subsubsection{Random Forest}
% Random Forest is classifier and regression method based on multiple decision trees that enjoy fast training and prediction which works well with high dimensional data. Thus it is a suitable candidate for handling large data-set  \cite{de_sousa_applying_2019, pandey_sele_2021}. 
%For instance, \cite{de_sousa_applying_2019} explores a  TDOA-based fingerprinting in the presence of the multi-path effect applying Random Forest. To capture the multi-path impact directly from the received signal (without performing the channel impulse response estimation), the volume cross-correlation (VCC) function takes two signals as input. It outputs a time delay estimation of the combination of their multi-path components. Furthermore, five peaks of VCC output (delay and amplitude) instead of just the estimated time delay are incorporated to handle the multi-path effect better. 

\subsubsection{Deep Learning}

The capability of Neural Networks (NN) to extract the complex input-output relation is advantageous for fingerprinting. More notably, NN can be used to directly map the fingerprints to the Cartesian coordinates like \cite{wye_rss-based_2021}, which incorporates ANN, taking RSS as input and outputting the estimated position in 2D. Deep learning techniques are more powerful than traditional ML techniques.  The main advantage of deep learning is capable of handling great deals of  complex multi-dimensional data that are corrupted by noise \cite{burghal2020comprehensive}. Deep learning can also make better use of the parallelism of GPU architecture, which is equivalent to lower run-time. There are, however, some challenges and disadvantages to deep learning models. First, they require the availability of rich, varied data representative of the problem to learn meaningful relationships between the input and output. Nevertheless, we lack tools to interpret models and understand for which part of the input space this relationship has been correctly learned and where there is a need for more data. This issue is also related to over-fitting, which appears when the network cannot generalize to unobserved data and makes improper predictions based on the training data. Lastly, finding suitable training hyper-parameters is a time-consuming practice.
% \begin{itemize}
%     \item lack of clear physical relationship between the data and results.  
%     \item availability of rich, varied data that is representative of the problem
%     \item over-fitting issue which making the trained network not to be able to be generalize to the unobserved data
%     \item finding the suitable model and hyper-parameters for the problem
% \end{itemize}
% }

One of the promising deep learning-based methods is CNN which has been widely deployed and used successfully for image classification. CNNs in online mode can perform very fast, though they require a meticulous and time-consuming training phase which still requires further investigation for fingerprinting localization. In \cite{sinha_comparison_2019}, a six-layer CNN classifier is used to learn and predict under the 74 classes. In \cite{song_novel_2019}, Stacked Auto-encoder (SAE) is incorporated to reduce the data dimension before feeding them into the CNN classifier for multi-floor multi-building localization.  In a 5G Internet of Things (IoT) test-bed, \cite{el_boudani_implementing_2020} combines two CNN for localization. The first one act as a regressor that estimates the position in 2D, whose output will be the input for the second neural network performing as a classifier to estimate the 3D location. Siamese network consisting of two sub-network (convolutional neural network with shared weights) with offline fine-tuning is proposed in \cite{pandey_sele_2021} to counter temporal changes of RSS, device heterogeneity and low RSS samples in a lightweight algorithm. In the training phase, an embedding function is learned that uniquely translates the distance and RSS pairs to an embedding. The training is executed so that each sub-network can take an RSS vector as input and estimate the relative distance of the locations corresponding to the RSS pairs. In the online step, the location is calculated as the weighting average of the RPs, based on the probabilities assigned for the embeddings of online RSS using Random Forest.

It is to be noted that these methods can also be combined in a multi-stage process for localization. For example, practically, the first simple rough estimation is done to find the sub-area candidate for the target position. Then a more accurate estimate is delivered by searching only through the selected sub-area. The work of \cite{nagy_rss-based_2020} clusters all data on the database based on signal levels involving RSS and directional antenna gain into the database. First, finding the initial rough estimation, the solution is refined with up-sampling until it meets the desired accuracy. The same two-step procedure is done in \cite{tan_uav_2020} for UAV positioning in 3D based on TOA. After coarse localization, neural network fitting is executed to refine the estimation. Finally, in \cite{he_hybrid_2020}, rough localization is performed based on TDOA, and the subarea is searched using a deep neural network (DNN) based on RSS. 

% \cite{he_hybrid_2020} exploiting RSS and TDOA extracted based on the uplink sounding reference signal (SRS) in LTE. Based on TDOA a coarse localization is performed thereafter the subarea is searched using a deep neural network (DNN) based on RSS. 

 % {\color{blue} accuracy, deployment difficulty, weakness, advantages, cost, reusability, robustness, power consumption\\
 
 % }

%  Parameters
% Additional Hardware
% Level of Accuracy
% Power Consumption
% Deployment
% Robustness
% Range Based Technique
% Needed
% High (85%-90%)
% High
% Hard
% High
% Range Free Technique
% Not Needed
% Low (70%-75%)
% Low
% Easy
% Low

%% file: Sections/section6.tex
\section{Other Taxonomies}
Aside from the techniques based on which the works fall under two broad main categories, there are other points of view that RF-based localization can be taken into account.

\subsection{Distributed vs Centralized }
Two approaches can be distinguished based on how calculation and process are done. In a centralized fashion, all data is sent to a central node and station where the computation is carried out. While this requires a less complex algorithm than the distributed approach, it requires a powerful processor. In a distributed way, the process and calculations are distributed among all nodes and subsystems, each node contributing to the final result. This method demands more adept and complex algorithms to be developed. As the number of nodes in WSN grows, distributed processes become more fascinating.

\subsection{Cooperative vs Non-cooperative}
Cooperatively, nodes communicate, share, and use the information to/from their neighbouring node. The advantage of the cooperation between nodes has been discussed in terms of Cramer-Rao Lower Bound (CRLB) \cite{schloemann_value_2016}. On the other hand, non-cooperative methods are more energy-efficient and less complex.

\subsection{Anchor-based vs Anchor-free}
In an anchor-based scheme, we have information about the location of some nodes or BS called anchors by either GPS or manually deploying them. This information and measurement are used as input for the localization algorithm. In the anchor-free method, however, there is no information about the positions of the nodes. While anchor-based are more accurate, anchor-free approaches are advantageous because they are more scalable and can remove the process of anchor deployment \cite{nazir_classification_2012, wen_decentralized_2008}.

\subsection{Static vs Mobile}
The majority of localization algorithms based on RF focused on the static target. The extension of these results to mobile targets, which is mostly the case in robotic applications, is non-trivial. One critical issue in mobile target localization is real-time feasibility. In the next section, we will review the localization of mobile robots. 

\subsection{Technologies}
Depending on the available infrastructure, desired accuracy, cost, and the environment, different technologies, such as WiFi, Bluetooth, Ultra-Wideband (UWB), Zigbee, Radio Frequency Identification Device (RFID), cellular network, Long Range (LoRA) radio, can be exploited. For example, UWB transmits signals across wide bandwidth in short range, delivering centimetre accuracy, and is suitable for indoor localization. On the other hand, WiFi benefits from exploiting existing WiFi infrastructure for localization. However, lower accuracy will be expected. More details can be found in \cite{zafari_survey_2019}.

\subsection{2D vs 3D}
A large part of the studies of localization so far have been addressed in 2D space. Theoretically, it is claimed that most of them can be extended to the 3D case, but in practice, localization in the vertical axis comes with a much higher error than the x-y axis. Therefore, 3D localization is worth further researching, which is highly important for drone applications. This issue is discussed more deeply in~\cite{kumari_localization_2019,coluccia_ml_2010}.

\subsection{Performance Parameters}

Evaluating the performance of the localization algorithm is not straightforward. Many criteria need to be considered for comparison: accuracy, precision, complexity, scalability, security, reliability, cost, and stability \cite{maghdid_comprehensive_2021}.

%% file: Sections/section7.tex
\section{ RF-based Localization for Aerial and Ground Robots}\label{rob_app}

%%%% REVIEEW 2
% 6. In the first paragraph of Section 7, the authors mentioned that “It is difficult to extend the current state-of-the-art for RF-based localization in robotic applications.” However, range-based and range-free algorithms have widely used in UAV and UGV applications. The authors should explain the difference between robotic applications and these traditional UAV/UGV applications and the reasons why the existing algorithms can’t be extended to robotic applications and what the challenges are.

% 7. In Section 7, the authors listed a lot of RF-based localization in robotic applications. However, the authors didn’t explain the specific meaning of these applications. The authors are suggested to target specific robotic applications when summarizing literatures.

 Most of the RF-based localization is applied to sensor network systems. What makes these attempts non-applicable to mobile vehicles in a straightforward way are the new challenges that the localization of UAVs and UGVs will bring.
\begin{itemize}
    \item UGVs and especially UAVs, are highly maneuverable with high speed. The existing state-of-the-art for WSN localization focuses on fixed targets and cannot address the fast changes in the target location and the real-time implementation. 
    \item The mobility of vehicles calls for a combination of other sensors like IMU and Images. The combination of the sensor data, especially images and RF, has not been studied in localization. 
    \item The majority of current works in WSN often consider just 2D cases, while vertical estimation is of great importance in UAV localization.
    \item The accuracy and robustness in demand in UAVs and UGVs localization applications are more critical. Usually, very accurate estimation is necessary, while in WSN, rather rough estimation suffices. This, for instance, rules out relying merely on RSS, which is the case for most of the existing state-of-the-art RF localization. 
    \item Use of limited technologies is the other drawback. For robot applications, UWB is mainly used, which is limited to indoors and is suitable for short range. New Technologies, especially 5G NR rarely been considered so far. In 5G, RSS would not be the most relevant feature, so there would be a shift to the use of this technology's new potentials and capabilities. 
\end{itemize}

Bayesian filters are among the most used methods for mobile robots localization, Unscented Kalman Filter \cite{sun_sensor_2010, you2020data}, Particle Filter \cite{wu_particle-filter-based_2014, luo_dynamic_2019, wang_dynamic_2021}, Extended Kalman Filter \cite{goel2018indoor,sung2008tdoa,lee2009tdoa, li2018accurate} and different variants of it like Square Root Cubature Kalman Filter (SRCKF) \cite{zhang_sequential_2016} and Diffusion Extended Kalman Filter (DEKF) \cite{xu2016distributed}.

% \cite{you2020data} fuses UWB and IMU with UKF for a quad-rotor UAV for both position and attitude estimation. Through actual implementation, centimetre-level accuracy is obtained.
Localization of a UGV moving with constant speed is developed in \cite{sung2008tdoa} using TDOA, incorporating an EKF with adaptive fading factor on updating the prediction covariance to account for the divergence issue of EKF. In a specific scenario \cite{goel2018indoor}, localization is performed for a group of 5 UAVs, one inside the room and the other four outdoors, based on the information gathered from GNSS, IMU, camera, UWB, and WiFi measurement by all the drones. UWB provides range measurements, and WiFi feeds RSS for indoor UAVs. EKF is used to fuse the measurements and jointly update and estimate all UAVs states and covariance considering the 6-DOF (degree-of-freedom) model. An event-driven sampling and transmission mechanism to counteract the effect of RSS volatility is proposed in \cite{zhang_sequential_2016}, where the anchors respond to the mobile robot if some conditions on the received signal are met. The work of \cite{lee2009tdoa} combines two EKFs to work simultaneously for location estimation based on TDOA. First, the model is augmented by a weighting filter used in one EKF framework to estimate the state. Then, this estimation is used by the other EKF to update the weights. In a decentralized strategy, \cite{xu2016distributed}, diffusion Kalman filter and EKF (DEKF) are integrated to estimate a target based on AOA where a group of UAVs based on AOA share their state estimation as well as Jacobian information.

In \cite{wu_particle-filter-based_2014}, PF is incorporated to estimate the position of a robot in 2D space based on RSS. Generating the radio map by collecting training fingerprints, Kernel Density Estimation is included to build the probabilistic observation likelihood for sample importance weights selection. The average of RSS at each point and the variance are recorded in the database. To enhance the accuracy, adaptive local search is employed to detect and remove unreasonable estimates by limiting the search area. Furthermore, a mechanism is applied to select just the subset of access points with lower variances to reduce the computation burden. The algorithm is tested for 4 cases with around 1 meter mean accuracy. The author in \cite{yucel_wi-fi_2020} fuses odometer data and WiFi RSS to track mobile robots using PF. Executing experiment in 2D, decimeter accuracy is delivered. A new particle filter-based algorithm is introduced in \cite{wang_dynamic_2021} to improve the estimation with fewer particles. To this goal, the traditional sequential prediction and update steps are done in one stage through the maximum likelihood estimator. Experimental results with WiFi RSS indoors show improvement, reaching about 0.5 $m^2$ means squared error. 

For target localisation, a fingerprinting method is explored in \cite{luo_dynamic_2019}. The radio map is continuously updated while collecting RSS and doing the target localization simultaneously. The database, the anchor positions, and propagation model parameters are stored and updated. Instead of using the log-distance model, a new model based on collected data is fitted. The model consists of three terms: path loss, measurement noise, and multi-path effect. Four simple functions are proposed and tested for the path loss model to select the best one: linear, nonlinear, and two log-distance models with different parameters. The result indicates that the log-distance model is not as accurate as the first two. The best model is of the form $c_1 d^{\lambda_1 d}+c_2e^{\lambda_2 d}$. Based on the model extracted, the fingerprints are modelled as the probability density function. In the sequel, a particle filter is employed for target tracking in online mode. %Comparing the RSS with the database, if any changes are detected either in anchor position or obstacle the shadowing model is remodelled, and the new beacon position is estimated. 

Two complex Maximum Likelihood (ML) algorithms are reported in \cite{li_enhanced_2021, cheng_communication-efficient_2022}, where simulations are conducted for evaluation. The work in \cite{li_enhanced_2021} utilizes Received Signal Strength (RSS) from multiple base stations (BS) along with trajectory information (velocity) of a drone to localize a UAV with unknown transmission power. A joint ML problem is formulated and solved based on the trajectory data and multiple BS. First, the positions in the ML function are fixed, and the transmission powers, assumed to be constant, are estimated. These estimates are then used to optimize the ML function for position estimation, using an exhaustive grid search. Two low-complexity alternative algorithms are also proposed, in which the original ML problem is decomposed into smaller, separately solved subproblems. The final result is calculated as a weighted combination of these separate estimations. Notably, the algorithms assume constant velocity at each time step.

A different scenario is considered in \cite{cheng_communication-efficient_2022}, where multi-UAVs are supposed to localize a fixed passive RF emitter based on the RSS. A group of UAVs flies along a predefined trajectory, with one UAV designated as the center of the formation. The authors propose a Distributed Majorization-Minimization method for RSS-based localization. To accelerate convergence, a tight upper bound of the primary objective function is derived, reducing the number of iterations required. A second-order Taylor expansion is employed to introduce a surrogate function, which is then optimized using the Majorize-Minimization algorithm. This process iterates through two steps: first, each UAV locally updates its position estimate based on the center UAV's estimate. Then, in the fusion step, the center UAV collects all local estimates and fuses them to generate a new global estimate, and distributes the updated result to the other UAVs. To reduce communication overhead, an alternative approach is proposed that requires only one round of communication, where edge UAVs transmit their local estimates to the center UAV. The center UAV then linearly fuses these estimates, using weights approximated from the Fisher information matrix. Simulation results show that this algorithm improves performance in terms of root mean square error (RMSE), although with higher computational complexity.

Another Maximum Likelihood Estimation (MLE)-based method is proposed in \cite{yang2021high}, which uses Ultra-Wideband (UWB) time-of-flight measurements. Non-convexity in the problem is addressed through linearization techniques. Additionally, the geometry of the anchor configuration and its impact on localization performance is examined using the Cramer-Rao Lower Bound (CRLB), with both simulations and experiments conducted to validate the approach.

Simultaneous Localization and Mapping (SLAM) using multiple robots is explored in \cite{liu_collaborative_2021}, leveraging WiFi fingerprinting and odometry in a 2D environment. A graph is constructed with nodes representing the robot poses, and constraints are derived from odometry, individual RSS fingerprints, and the similarity of RSS fingerprints between robots. For trajectory optimization using graph-based SLAM, the distance between fingerprints and their variance is derived from the database using a simple model.

In contrast to conventional deterministic propagation models, \cite{zickler_rss-based_2010} employs a probabilistic likelihood function aided by a symmetric trapezoidal distribution over discrete localization grids. Tethering (where one mobile node localizes, tracks, and follows another mobile node) is achieved by integrating RSS and odometry information. The 2D area is divided into grids with discrete probabilities, and a Bayesian update is used to revise the grid probabilities upon receiving a new RSS value. This approach allows for extracting the current location by obtaining the full posterior probability distribution over all grids.

Stojkoska et al.\cite{stojkoska_indoor_2017} incorporate Multi-dimensional Scaling (MDS) and Weighted Centroid Localization (WCL) for indoor localization. Al-Jazzar and Jaradat\cite{al2020aoa} propose a geometrical approach where the UAV’s position is determined through mathematical techniques based on six AOA measurements from sensor doublets. In \cite{xu2020three}, the Least Squares (LS) method is used for UAV localization in 3D by fusing AOA and TDOA data. Nguyen et al.~\cite{nguyen2019integrated} explore relative localization by estimating the UAV's position relative to a target, a method useful for formation control \cite{guo2020simultaneous} and autonomous docking. Their work combines Ultra-Wideband (UWB) range measurements with vision-based data, achieving decimeter-level accuracy in a 2D experimental setup using recursive LS. These methods are compared in \autoref{table0}. In a more recent study, \cite{meles2021measurement} investigates the performance of AOA for UAV localization in a cellular network, reporting an accuracy of less than 45 meters.

As noted, the current state-of-the-art in robot localization is limited to very specific scenarios, technologies (like UWB), RF features, and sensor data. Two key limitations must be addressed:

\begin{itemize}
    \item Limited to specific technologies and sensor data: Most papers use RSS due to easy hardware. In that case, acceptable accuracy is just achieved by using UWB which is limited for indoors with short range. TOA-based localization is also done mostly by taking advantage of UWB. Moreover, many possibilities are missing in the literature like the integration of images, and LIDAR with RFs. 
    
    \item Limited Accuracy: accuracy is one of the main concerns in UAVs and UGVs localization. Only relying on simple algorithms and sensor data, like RSS, might not be an appropriate solution, especially with the upcoming technologies, 5G and beyond. As we discuss later in our paper, CSI information would provide a huge amount of useful data. However, the real-time implementation and its fusion with conventional sensor data is the real concern that is not addressed. Edge computing and off-loading as the most promising solutions are rarely investigated.
    
\end{itemize}

There is a huge gap worth filling in this area: use of RF in SLAM, integrating variant of sensor data with RF, use of CSI in robot applications, and developing Deep learning approach, especially with the fingerprinting method.

 %%%%%%%%%%%%%%%%%%%%%%%%%%%%%%%%%%%%%%%%%%%%%%%%%%%%%%%%%%%%%%%%%%%%%%%%%%%%%%%%%%%%%%%
 \clearpage

\begin{landscape}
% \hskip -50.0cm 
% \hspace{-10cm}
\renewcommand\theadset{\def\arraystretch{0.5}}%
\renewcommand\theadfont{\fontsize{7}{5}\selectfont}
\begin{flushleft}
\begin{table*}[!t]
\centering

\caption{Comparisons between existing works on RF-based localization for UGVs and UAVs .}
\label{table0}
\scalebox{1}{

\begin{tabularx}{1.16	\textwidth}{m{1cm}m{0.5cm}m{ 2.1cm}m{ 1.3cm}m{2.2cm}m{1.5cm}m{ 0.5cm}m{ 1.5cm}m{ 1.21cm}m{3cm}XXXXXXXXXX}

\Xhline{3\arrayrulewidth}
\thead{\bf Refs} & \thead{\bf Year}& \thead{ \bf Range-based/ \\ \bf Fingerprinting} & \thead{  \bf Distributed/ \\ \bf Centralized} & \thead{ \bf Cooperative/ \\ \bf Non-cooperative} &  \thead{\bf Anchor-based/ \\ \bf Anchor-free} & \thead{\bf 2D/3D}  &\thead{\bf Experiment} & \thead{\bf Technique} & \thead{\bf Technology}\\ %anchorbased anchor free
%------------------------------------------------------------------------------------------------------------------------------------
\Xhline{3\arrayrulewidth}

%------------------------------------------------------------------------------------------------------------------------------------
% \Xhline{2\arrayrulewidth}
\cellcolor[gray]{.94}\thead{\cite{wu_particle-filter-based_2014}} & \cellcolor[gray]{.94}\thead{2014}& \cellcolor[gray]{.94}\thead{fingerprinting} & \cellcolor[gray]{.94}\thead{centralized} & \cellcolor[gray]{.94}\thead{non-cooperative}  & \cellcolor[gray]{.94}\thead{anchor-free} & \cellcolor[gray]{.94}\thead{2D} & \cellcolor[gray]{.94}\thead{Yes} & \cellcolor[gray]{.94}\thead{PF}& \cellcolor[gray]{.94}\thead{WLAN/RSS} \\ %\hline
%------------------------------------------------------------------------------------------------------------------------------------
% \Xhline{2\arrayrulewidth}
\cellcolor[gray]{1} \thead{\cite{sun_sensor_2010}} &\cellcolor[gray]{1}\thead{2010}& \cellcolor[gray]{1}\thead{range-based} & \cellcolor[gray]{1}\thead{centralized}  & \cellcolor[gray]{1}\thead{non-cooperative} & \cellcolor[gray]{1}\thead{anchor-based}  & \cellcolor[gray]{1}\thead{2D} & \cellcolor[gray]{1}\thead{Yes} & \cellcolor[gray]{1}\thead{UKF}& \cellcolor[gray]{1}\thead{-/RSS} \\
% This mentioned wihtou requiring any reference position maybe all fingerprinting are anchor-free

%------------------------------------------------------------------------------------------------------------------------------------
% \Xhline{2\arrayrulewidth}
\cellcolor[gray]{.94}\thead{\cite{you2020data}} &\cellcolor[gray]{.94}\thead{2020}& \cellcolor[gray]{.94}\thead{range-based} & \cellcolor[gray]{.94}\thead{centralized} & \cellcolor[gray]{.94}\thead{non-cooperative}  & \cellcolor[gray]{.94}\thead{anchor-based} & \cellcolor[gray]{.94}\thead{3D} & \cellcolor[gray]{.94}\thead{Yes} & \cellcolor[gray]{.94}\thead{UKF}& \cellcolor[gray]{.94}\thead{UWB/TOA} \\ %\hline
%------------------------------------------------------------------------------------------------------------------------------------
% \Xhline{2\arrayrulewidth}
\cellcolor[gray]{1} \thead{\cite{luo_dynamic_2019}} &\cellcolor[gray]{1}\thead{2019}& \cellcolor[gray]{1}\thead{fingerprinting} & \cellcolor[gray]{1}\thead{centralized}  & \cellcolor[gray]{1}\thead{non-cooperative} & \cellcolor[gray]{1}\thead{anchor-based}  & \cellcolor[gray]{1}\thead{2D} & \cellcolor[gray]{1}\thead{Yes} & \cellcolor[gray]{1}\thead{PF}& \cellcolor[gray]{1}\thead{ZigBee/RSS} \\
%------------------------------------------------------------------------------------------------------------------------------------
% \Xhline{2\arrayrulewidth}
\cellcolor[gray]{.94}\thead{\cite{wang_dynamic_2021}} &\cellcolor[gray]{.94}\thead{2021}& \cellcolor[gray]{.94}\thead{fingerprinting} & \cellcolor[gray]{.94}\thead{centralized} & \cellcolor[gray]{.94}\thead{non-cooperative}  & \cellcolor[gray]{.94}\thead{anchor-free} & \cellcolor[gray]{.94}\thead{2D} & \cellcolor[gray]{.94}\thead{Yes} & \cellcolor[gray]{.94}\thead{PF-ML}& \cellcolor[gray]{.94}\thead{WiFi/RSS} \\ %\hline
%------------------------------------------------------------------------------------------------------------------------------------
 \thead{\cite{goel2018indoor}} &\cellcolor[gray]{1}\thead{2018}& \thead{range-based} & \thead{centralized}  & \thead{cooperative} & \thead{anchor-based}  & \thead{3D} & \thead{No} & \thead{EKF}& \thead{WiFi-UWB/RSS-TOA} \\
%------------------------------------------------------------------------------------------------------------------------------------
\cellcolor[gray]{.94}\thead{\cite{sung2008tdoa}} &\cellcolor[gray]{.94}\thead{2008}& \cellcolor[gray]{.94}\thead{range-based} & \cellcolor[gray]{.94}\thead{centralized} & \cellcolor[gray]{.94}\thead{non-cooperative}  & \cellcolor[gray]{.94}\thead{anchor-based} & \cellcolor[gray]{.94}\thead{2D} & \cellcolor[gray]{.94}\thead{No} & \cellcolor[gray]{.94}\thead{EKF}& \cellcolor[gray]{.94}\thead{-/TDOA} \\ %\hline
%------------------------------------------------------------------------------------------------------------------------------------

% \Xhline{2\arrayrulewidth}
\thead{\cite{lee2009tdoa}} &\thead{2009}& \thead{range-based} & \thead{centralized} & \thead{non-cooperative}  & \thead{anchor-based} & \thead{2D} & \thead{No} & \thead{EKF}& \thead{-/TDOA} \\ %\hline
%------------------------------------------------------------------------------------------------------------------------------------
% \Xhline{2\arrayrulewidth}
\cellcolor[gray]{.94} \thead{\cite{li2018accurate}} &\cellcolor[gray]{0.94}\thead{2018} &\cellcolor[gray]{.94}\thead{range-based} & \cellcolor[gray]{.94}\thead{centralized}  & \cellcolor[gray]{.94}\thead{non-cooperative} & \cellcolor[gray]{.94}\thead{anchor-based}  & \cellcolor[gray]{.94}\thead{3D} & \cellcolor[gray]{.94}\thead{Yes} & \cellcolor[gray]{.94}\thead{LS}& \cellcolor[gray]{.94}\thead{UWB/TOA} \\
%------------------------------------------------------------------------------------------------------------------------------------

% \Xhline{2\arrayrulewidth}
\cellcolor[gray]{1}\thead{\cite{zhang_sequential_2016}} & \cellcolor[gray]{1}\thead{2016}&\cellcolor[gray]{1}\thead{range-based} & \cellcolor[gray]{1}\thead{centralized} & \cellcolor[gray]{1}\thead{non-cooperative}  & \cellcolor[gray]{1}\thead{anchor-based} & \cellcolor[gray]{1}\thead{2D} & \cellcolor[gray]{1}\thead{Yes} & \cellcolor[gray]{1}\thead{SRCKF}& \cellcolor[gray]{1}\thead{-/RSS} \\ %\hline

%------------------------------------------------------------------------------------------------------------------------------------
% \Xhline{2\arrayrulewidth}
\cellcolor[gray]{.94}\thead{\cite{xu2016distributed}} &\cellcolor[gray]{.94}\cellcolor[gray]{0.94}\thead{2016}& \cellcolor[gray]{.94}\thead{range-based} & \cellcolor[gray]{.94}\thead{distributed} & \cellcolor[gray]{.94}\thead{cooperative}  & \cellcolor[gray]{.94}\thead{anchor-based} & \cellcolor[gray]{.94}\thead{2D} & \cellcolor[gray]{.94}\thead{No} & \cellcolor[gray]{.94}\thead{DEKF}& \cellcolor[gray]{.94}\thead{-/AOA} \\ %\hline

%------------------------------------------------------------------------------------------------------------------------------------
% \Xhline{2\arrayrulewidth}
\cellcolor[gray]{1}\thead{\cite{yucel_wi-fi_2020}} &\cellcolor[gray]{1}\thead{2020}& \cellcolor[gray]{1}\thead{fingerprinting} & \cellcolor[gray]{1}\thead{centralized} & \cellcolor[gray]{1}\thead{non-cooperative}  & \cellcolor[gray]{1}\thead{anchor-free} & \cellcolor[gray]{1}\thead{2D} & \cellcolor[gray]{1}\thead{Yes} & \cellcolor[gray]{1}\thead{PF}& \cellcolor[gray]{1}\thead{-/RSS} \\ %\hline
%------------------------------------------------------------------------------------------------------------------------------------
% \Xhline{2\arrayrulewidth}
\cellcolor[gray]{.94}\thead{\cite{li_enhanced_2021}} &\cellcolor[gray]{.94}\thead{2021}& \cellcolor[gray]{.94}\thead{range-based} & \cellcolor[gray]{.94}\thead{centralized} & \cellcolor[gray]{.94}\thead{non-cooperative}  & \cellcolor[gray]{.94}\thead{anchor-based} & \cellcolor[gray]{.94}\thead{3D} & \cellcolor[gray]{.94}\thead{No} & \cellcolor[gray]{.94}\thead{ML}& \cellcolor[gray]{.94}\thead{-/RSS} \\
%------------------------------------------------------------------------------------------------------------------------------------
% \Xhline{2\arrayrulewidth}
\cellcolor[gray]{1}\thead{\cite{cheng_communication-efficient_2022}}  &\cellcolor[gray]{1}\thead{2022}& \cellcolor[gray]{1}\thead{range-based} & \cellcolor[gray]{1}\thead{distributed} & \cellcolor[gray]{1}\thead{cooperative}  & \cellcolor[gray]{1}\thead{anchor-based} & \cellcolor[gray]{1}\thead{3D} & \cellcolor[gray]{1}\thead{No} & \cellcolor[gray]{1}\thead{ML}& \cellcolor[gray]{1}\thead{-/RSS} \\ %\hline
%------------------------------------------------------------------------------------------------------------------------------------
%\toprule[1pt] \\ %[1cm] %need booktabs
% \Xhline{2\arrayrulewidth}
\cellcolor[gray]{.94} \thead{\cite{yang2021high}} &\cellcolor[gray]{0.94}\thead{2021}& \cellcolor[gray]{.94}\thead{range-based} & \cellcolor[gray]{.94}\thead{centralized}  & \cellcolor[gray]{.94}\thead{non-cooperative} & \cellcolor[gray]{.94}\thead{anchor-based}  & \cellcolor[gray]{.94}\thead{3D} & \cellcolor[gray]{.94}\thead{Yes} & \cellcolor[gray]{.94}\thead{EKF}& \cellcolor[gray]{.94}\thead{UWB/TW-TOF} \\
%------------------------------------------------------------------------------------------------------------------------------------
% \Xhline{2\arrayrulewidth}
\cellcolor[gray]{1} \thead{\cite{liu_collaborative_2021}} &\cellcolor[gray]{1}\thead{2021} &\cellcolor[gray]{1}\thead{fingerprinting} & \cellcolor[gray]{1}\thead{centralized}  & \cellcolor[gray]{1}\thead{cooperative} & \cellcolor[gray]{1}\thead{anchor-free}  & \cellcolor[gray]{1}\thead{2D} & \cellcolor[gray]{1}\thead{Yes} & \cellcolor[gray]{1}\thead{ML}& \cellcolor[gray]{1}\thead{WiFi/RSS} \\

%------------------------------------------------------------------------------------------------------------------------------------
% \Xhline{2\arrayrulewidth}
\cellcolor[gray]{.94} \thead{\cite{zickler_rss-based_2010}} &\cellcolor[gray]{0.94}\thead{2010}& \cellcolor[gray]{.94}\thead{range-based} & \cellcolor[gray]{.94}\thead{centralized}  & \cellcolor[gray]{.94}\thead{non-cooperative} & \cellcolor[gray]{.94}\thead{anchor-based}  & \cellcolor[gray]{.94}\thead{2D} & \cellcolor[gray]{.94}\thead{Yes} & \cellcolor[gray]{.94}\thead{Bayesian}& \cellcolor[gray]{.94}\thead{-/RSS} \\
% This mentioned wihtou requiring any reference position maybe all fingerprinting are anchor-free

%------------------------------------------------------------------------------------------------------------------------------------

% \Xhline{2\arrayrulewidth}
\cellcolor[gray]{1} \thead{\cite{meles2021measurement}} &\cellcolor[gray]{1}\thead{2021}& \cellcolor[gray]{1}\thead{range-based} & \cellcolor[gray]{1}\thead{centralized}  & \cellcolor[gray]{1}\thead{non-cooperative} & \cellcolor[gray]{1}\thead{anchor-based}  & \cellcolor[gray]{1}\thead{2D} & \cellcolor[gray]{1}\thead{Yes} & \cellcolor[gray]{1}\thead{LS}& \cellcolor[gray]{1}\thead{Cellular/AOA} \\
%------------------------------------------------------------------------------------------------------------------------------------
% \Xhline{2\arrayrulewidth}
\cellcolor[gray]{.94} \thead{\cite{stojkoska_indoor_2017}} &\cellcolor[gray]{0.94}\thead{2017}& \cellcolor[gray]{.94}\thead{range-based} & \cellcolor[gray]{.94}\thead{centralized}  & \cellcolor[gray]{.94}\thead{non-cooperative} & \cellcolor[gray]{.94}\thead{anchor-based}  & \cellcolor[gray]{.94}\thead{3D} & \cellcolor[gray]{.94}\thead{No} & \cellcolor[gray]{.94}\thead{MDS-WCL}& \cellcolor[gray]{.94}\thead{WiFi/RSS} \\
%------------------------------------------------------------------------------------------------------------------------------------
% \Xhline{2\arrayrulewidth}
\thead{\cite{al2020aoa}} &\thead{2020} &\thead{range-based} & \thead{centralized} & \thead{non-cooperative}  & \thead{anchor-based} & \thead{3D} & \thead{No} & \thead{Lateration}& \thead{-/AOA} \\ %\hline
%------------------------------------------------------------------------------------------------------------------------------------
% \Xhline{2\arrayrulewidth}
\cellcolor[gray]{.94} \thead{\cite{xu2020three}} &\cellcolor[gray]{.94}\thead{2020}& \cellcolor[gray]{.94}\thead{range-based} & \cellcolor[gray]{.94}\thead{centralized}  & \cellcolor[gray]{.94}\thead{non-cooperative} & \cellcolor[gray]{.94}\thead{anchor-based}  & \cellcolor[gray]{.94}\thead{3D} & \cellcolor[gray]{.94}\thead{No} & \cellcolor[gray]{.94}\thead{LS}& \cellcolor[gray]{.94}\thead{-/TDOA-AOA} \\
%\hline

%------------------------------------------------------------------------------------------------------------------------------------
% \Xhline{2\arrayrulewidth}
\cellcolor[gray]{1} \thead{\cite{nguyen2019integrated}} &\cellcolor[gray]{1}\thead{2019}& \cellcolor[gray]{1}\thead{range-based} & \cellcolor[gray]{1}\thead{centralized}  & \cellcolor[gray]{1}\thead{non-cooperative} & \cellcolor[gray]{1}\thead{anchor-based}  & \cellcolor[gray]{1}\thead{2D} & \cellcolor[gray]{1}\thead{Yes} & \cellcolor[gray]{1}\thead{RLS}& \cellcolor[gray]{1}\thead{UWB/TOA} \\

 %\hline
%\toprule[1pt] \\ %[1cm] %need booktabs

\Xhline{3\arrayrulewidth}
 \end{tabularx}
 }
%    {\hspace{3ex} \raggedright \fontsize{8}{12} \selectfont 1: Range-based/Fingerprinting (Rb/F); 2:  Centralized/Distributed (cent/dist); 3: collaborative/non-collaborative(col/n-col) ; 4: Anchor-based/ Anchor-free (Ab/Af); 5: 2D/3D 6: Real Experiment; 7: Technique; 8: Technology  \\}

% }
 \end{table*}
 \end{flushleft}
 \end{landscape}

%%%%%%%%%%%%%%%%%%%%%%%%%%%%%%%%%%%%%%%%%%%%%%%%%%%%%%%%%%%%%%%%%%%

%% file: Sections/section8.tex
\section{5G potentials and promises for robot applications}

Rolling out of the 5G New Radio (NR) technology provides great potential to boost the localization of robots and UAVs in terms of accuracy, robustness, cost, and coverage. The promising features of 5G NR for robot applications include the following:
\begin{figure}
\caption{5G NR enabler for improved robot localization}
\centering
\includegraphics[width = 0.5 \textwidth]{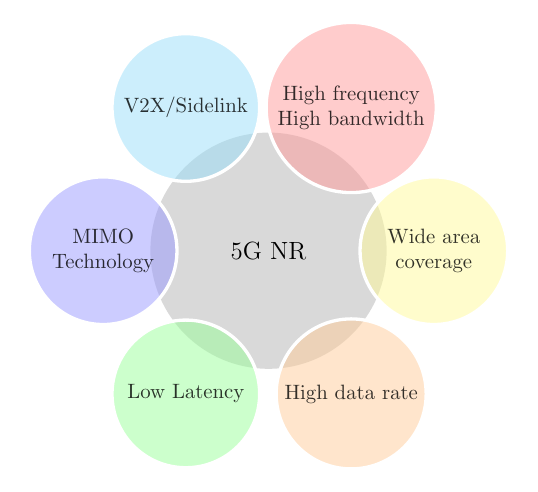}
\end{figure}

\begin{itemize}
\item Wide area coverage
\item MIMO technology
\item High carrier frequency
\item High bandwidth
\item Vehicle-to-Everything (V2X)
\item Low latency
\item High throughput
\end{itemize}

\subsection{Wide area coverage and in-expensive localization systems}
Compared to other technologies like WiFi, UWB, etc., 5G will be available almost anywhere, indoors or outdoors, since the cellular infrastructure is widely deployed in the cities. Using Vehicle-to-Everything (V2X) also makes it feasible to take advantage of 5G, where there is no full coverage. For example, in a collaborative scheme, one or some part of the device(s) or vehicle(s) can play the role of anchor or Pseudo BS for others. 5G is also considered an in-expensive solution because there is no specific equipment to set up as long as operating under the coverage of BS. Accordingly, taking advantage of the available infrastructure of 5G, in some cases, might remove the need for costly and also energy-consuming GPS devices. 

In addition, 5G confers robustness to the localization system. Robustness is an essential feature in highly mobile scenarios, in the case where safety is of great concern. For example, consider a vehicle or a group operating in a wide area where some part of it might be GPS-denied. In such a situation, relying only on GPS information may cause failure. 

% (does not fit well)? It is mentioned that each RF technique has its own disadvantages and weaknesses. TOA and TDOA require very tight time synchronization. Just 1 $\mu$s might lead to more than 300 meters ranging estimation error. The performance of the angle-based method degrades when there is no LOS, and the target is far from the BS. So exploiting a hybrid method (combination of diverse measurements) based on the situation can result in more accurate and robust localization. % The data that might include the odometry information provided by IMU/camera, GPS, RSS, AOA, TOA, and TDOA.

\subsection{RF Measurements with more resolution}
The 3GPP 5G New Radio (NR) is envisioned to pave the way for achieving higher accuracy and robustness of localization in GPS-denied environments like indoors and improving localization outdoors combined with GPS. 5G NR provides improved measurements for localization like time-based, angular-based, and energy-based measurements. The measurements include TOA, TDOA, AOA/AOD, and multi-cell round-trip time. For robot localization, 5G in the downlink defines a new reference signal called positioning reference signal (PRS) based on which these measurements can be extracted. (the readers are referred to \cite{dwivedi_positioning_2021} for further information ).

Owing to the high carrier frequency, high bandwidth, and MIMO technology accurate measurements will be delivered. 5G NR operates at high-frequency bands: Frequency Range 1 (FR1) (450 MHz to 6 GHz) and Frequency Range 2 (FR2) ( frequency bands from 24.25 GHz to 52.6 GHz). Relying on the Cramer-Rao Lower Bound (CRLB), the lower bound for variance of TOA is obtained by \cite{zhang2017cooperative}:
\begin{gather*}
    var(TOA)\ge \frac{1}{8\pi^2BT_sF_c^2 \textsc{ SNR}} \,,
\end{gather*}
with B being the bandwidth, $F_c$ is the central frequency, $T_s$ is the duration of the signal, and SNR is signal to noise ratio. This inequality indicates that higher frequency and bandwidth contribute to higher TOA estimation accuracy. 

 The probability of LOS increases due to the strong path loss in higher frequencies (mmWave). This feature and the large transmission bandwidth make distinguishing between LOS and NLOS measurements in the multipath effect feasible by applying proper analysis over the received signals, resulting in a highly precise time-based and angular-based localization.  

The Massive MIMO technology is one of the most noticeable enhancements and relevant features for localization offered by 5G NR. It allows for the implementation of ultra-massive antenna arrays consisting of hundreds or thousands of antennas in a single base station leading to the finer angular resolution (azimuth and elevation of the beam) of even less than one degree which can contribute to accurate localization.

\subsection{Vehicle-to-Everything Standard}

The Third Generation Partnership Project (3GPP) deploys the Vehicle-to-Everything (V2X) based on Dedicated Short Range Communications which includes Vehicle-to-vehicle (V2V), Vehicle-to-Network (V2N) or Vehicle-to-Infrastructure (V2I). The cellular V2X standard based on the 5G air interface is a fascinating feature for cooperative robotic missions, specifically through introducing the sidelink (SL), which permits vehicles to directly exchange information without other parts of the network being involved. This will play a vital role in cooperative tasks and localization, whether the operation is within the coverage area or without BS coverage. Furthermore, the V2X feature allows for not only communication of the vehicles with each other but with the infrastructure and even the internet. This increased connectivity improves the efficiency of the cooperative systems, enhances localization accuracy, and makes some missions feasible which were not before. For example, in an environment with adverse NLOS effects from the BS, which results in erroneous measurements, one or some parts of the vehicles can play the role of BSs or act as a pseudo BS.

\subsection{Low Latency}

Ultra-reliable low latency communication (URLL) offered in 5G new radio allows for future applications which is on-demand for aggressive latency for quick reaction. The existing 4G cellular network is not appropriate for this purpose.

%multi-robot systems which entails a real-time high transmission rate.

 Low latency means a slight delay between sending and receiving information indispensable for autonomous driving and flying. For example, 1 ms minimum latency (average of 10 ms) is expected to be brought by 5G, which is a substantial breakthrough compared to the 200 ms latency typical of 4G. The central station that might do the localization, planning, or control must receive and send back data and command fast enough to the vehicles operating at high speed. The time for transmitting and receiving data is vital for autonomous control, where the car moves rapidly. For example, for a drone to be controlled, both localization and control commands need to be processed and sent back to the UAV with an acceptable slight delay, resulting in a fast reaction of the UAV. 
 
 % \cite{kutila20205g} some experiments on latency over 5g and bandwidth
 
 %This feature is also useful for the cooperation scenario where robots or drones need to share their data with others to do some cooperative tasks like coverage, surveillance, or collaborative rescue mission.

 \subsection{High throughput}

Significantly for the uplink transmission, a high data rate is needed when offloading computations. Many algorithms try to balance computation power and accuracy mostly because onboard computers are usually not equipped with a powerful processing unit. This concern would be prevented with offloading, which allows for implementing complex algorithms with high accuracy in real-time on a powerful server using edge computing. In addition to low latency, a high data rate is crucial for the real-time transfer of extensive sensor data -such as high-resolution images or LIDAR data. In \cite{hayat2021edge}, the role of edge computing and the impact of data rate in the uplink for vision-based drone navigation in the 5G network is explored. Three scenarios are tested and compared: no offloading, partial, and complete offloading. It shows how offloading can be advantageous in the network capable of providing a fast uplink rate.

% \subsection{Multi-robot SLAM}

% A critical task in robotics is Simultaneous Localization and Mapping in which as robot moves in an unfamiliar environment it needs to construct a map and localize itself on the map while navigating. The use of multi robots instead of a single one, while each shares its maps with others, is a fascinating topic that adds efficiency to the SLAM.  Multi-robot SLAM (MRSLAM) is implemented in two ways. Using a central station to which all robots disseminate the collected data to do all the processing and construct the global map and transmit back this data to each robot. In a decentralized system, each robot performs locally, and whenever visit other robots share their local map based on which they update their map. 
% Accordingly, one big challenge in both centralized and distributed MRSLAM is the constraint on the communication bandwidth, limited computation power, and memory. 5G NR will circumvent these limitations by facilitating edge computing \cite{huang2022edge, huang2021colaslam}, providing accurate RF-based measurements, and enabling relative localization for each vehicle in the team via sidelink and V2V technologies.

\subsection{Localization based on 5G}
%  This same property can also be harnessed to determine
 
%   -The methods developed for LOS 
 
%  -intermediate parameters such as AOA
%  -instead of explore, demonstrate can be used
 
%   -NLOS signals will cause large errors in TOA result 
 
%  -since 5G deployments are in the early stages with limited availability, it is challenging to conduct a large-scale comparative study between LTE and 5G
 
  As the 5G roll-out is in the early stages, not many works have investigated 5G localization, especially for robotics applications and in a sensor fusion framework. Besides, most are just based on simulation results, neglecting many practical aspects. For example, the impact of synchronization error in time-based positioning or simplification of channel models is considered. In the following, we review the current state-of-the-art 5G-based localization. 
 
 In 5G NR, pilot signals are included for positioning purposes, Positioning Reference Signal (PRS) in down-link and Sounded Reference Signal (SRS) in up-link \cite{dwivedi_positioning_2021}. In addition to network centre frequency and bandwidth, PRS and SRS configuration also play roles in localization accuracy.
 In \cite{ferre2019positioning}, for different combinations of centre frequency, sub-carrier spacing, and PRS comb-size, localization accuracy for simple scenarios is compared in terms of Root Mean Square Error (RMSE) using simulation. The authors in \cite{del2016feasibility} simulate the roadside 5G network implementation for assisted driving, showing accuracy below 20-25 cm for 50-100 MHz bandwidth. Localization is performed based on TOA extracted as the first correlation peak between PRS and the received signal. The channel is modelled based on path loss and the TDL channel model. The impact of the geometrical placement of roadside 5G base stations on the localization based on EKF and how the distance from BS affects EKF linearization error is investigated in \cite{saleh20225g}. In this paper, EKF for location estimation is also presented where the covariance matrix is tuned dynamically, and improvement is shown through simulation. There are several attempts for localization in the 5G network based on CIR and CFR. The work of \cite{panjoint} addresses localization in the up-link side based on 5G SRS information, in which, based on the received signal, the channel frequency response (CFR) is estimated. TOA and Direction of Arrival (DOA) are then evaluated using the well-known 2D multiple signal classification (MUSIC) algorithms. Localization is finally performed, and an indoor experimental setup achieves an accuracy of less than 1m. The authors in \cite{zhang2021aoa} generated a fingerprint dataset of AOA and its corresponding amplitude based on the CSI matrix in a 5G network. Deep Neural network (DNN) is trained and used as a regressor for online estimation. Quasi Deterministic Radio channel Generator (QuaDRiGa) \cite{jaeckel2014quadriga} is exploited for channel modelling. Approximately 1-meter accuracy for NLOS and 0.1 m for LOS are reported. Based on the SRS symbol, in up-link, CFR is estimated for each base station in \cite{panlow}, and subsequently, TOA and AOA are jointly estimated for localization. Accuracy below 1m is acquired. Localization under the fingerprinting framework is explored in \cite{stahlke2022transfer} based on CSI, where the transfer learning concept is leveraged to reduce the real-world training effort. QuaDRiGa is leveraged for obtaining synthetic CSI to pre-train the CNN model. In \cite{meng2020study}, angle-based fingerprint localization is conducted. The fingerprints include the angles (zenith and azimuth) along with their corresponding power for all observed paths. To validate the results, simulation is done by recreating 3D outdoor environments, including building geometry. AOA-based position estimation in 5G network experimentally reported in \cite{menta2019performance}, where EKF is used at edge cloud for localization. The research of \cite{sellami2020multi,klus2021neural} is related to receiver localization harnessing AOA under 5G MIMO System and beam-formed RSS, respectively. 

While TOA, TDOA, AOA, and RSS of the LOS path could be directly related to the relative positions of the transmitter and receiver, there is no explicit connection between the NLOS path and close distance. Thus, the localisation performance will be degraded noticeably only relying on those measurements. In literature, NLOS error mitigation techniques and ray tracing-based approaches are carried out to compensate for the NLOS ( see \cite{wen2019survey} and the references therein). Under the 5G network, this issue is addressed in \cite{deng2020tdoa}. The effect of NLOS in an unknown environment is dealt with in a fusion framework. Localization and navigation are accomplished by fusing TDOA and Pedestrian Dead Reckoning (PDR). TDOA from LOS base stations is combined with TDOA from virtual Base stations placed in an unknown area whose locations are determined based on the NLOS base stations. Simulation and experiments are performed. The work of \cite{mendrzik2018joint} goes beyond just localization by mapping the radio environment simultaneously, taking advantage of NLOS-rich information about transmitter and receiver positions and environmental obstacles. This paper proposes joint position and orientation estimation for a mobile target and the position estimation of reflectors and scatters relying on NLOS paths. Leveraging AOA, AOD, and TOA for each NLOS path, the receiver's location is determined only based on the received signal from one base station in 2D if there are at least 3 NLOS paths. It is shown in \cite{chu2021vehicle}, in NLOS situation, cooperation among vehicles improves situational awareness and localization performance as several cars operate in the same environment where they might share one or more scatters, which results in a correlated multipath structure that can contribute to the improved localization.

%% file: Sections/section9.tex
\section{Future Research Directions and Challenges}
%% REVIEWER COMMENTS

% Advantages and disadvantages of deep learning for different RF-based localization approaches should be presented briefly.

% In fact, the Section 9 is quite important and attracts significant attention of readers. So, it is better if authors can discuss further about the vertical localization accuracy and safety.

Most of the research on the use of 5G for localization approaches the problem from the pure communication point of view, while its use in various robotic applications is still in the infancy stage. In this section, we shed light on future opportunities, research gaps, and challenges that will be provided by 5G for UAV and UGV applications. 

\subsection {Fingerprinting and Deep Learning applied to CSI}
Compared to range-based methods, fingerprinting has more potential to deliver higher localization precision. On the one hand, range-based methods are limited by the accuracy of the model they apply to extract ranges or angles. On the other hand, fingerprinting may utilize the signal information to its full extent. Mainly, CSI data contains latent knowledge that can be captured by the complex AI approaches linked with fingerprinting. Hence, appropriate Deep learning models may be the key to exploiting the CSI large matrix structure effectively to hallucinate necessary abstract features for localization. For example, deep neural networks may be able to decode these high-dimensional matrices to localize the obstacles around the receiver from the NLOS path. 
 
Furthermore, if multi-array antennae are deployed at the receiver and transmitter, CSIs contain even richer information since different versions of the same signal would be available as separate fingerprints. Therefore, effectively profiting from CSI information may prevent the need for several BSs for localization. It is shown in \cite{mendrzik2018joint} that just one BS might suffice. Lastly, in this method, the main disadvantage of the traditional RF-based method, i.e., NLOS situations, is circumvented, under which these approaches become erroneous and unreliable. 

There are challenges in fingerprinting that are worth much more attention. The availability of data and the collection of enough data is not trivial. The collected data set is not entirely reliable as the environment constantly evolves. Transfer learning as one possible solution is suggested to exploit synthetic data or already available data-set. In this area, there still seems to be much room to investigate the integration of deep learning with CSI fingerprints. % combined with other sensor data in both theoretical and practical aspects. 
% are the best seems promising to handle effectively to learn complex hidden features encoded in the fingerprints for not just localization of the receiver but also estimation of the scatter positions, which adds to the situational awareness.
% CSI provide complete and informative information that can be used directly as fingerprints. 

% though the amount of information is vast, and the complexity is higher, leveraging Deep Learning seems promising to handle effectively learning complex hidden features encoded in the fingerprints for not just localization of the receiver but also estimation of the scatter positions, which adds to the situational awareness.

\subsection{Fusion of RF with other sensor data}
% Relying on CSI, even one base station might be enough to localize the target. Moreover, it makes the localization system robust against NLOS situations under which other methods based on the different features become erroneous. 
The mobility and the issue of receiving diverse information with different frequencies is a challenge that is missing in the literature. Localization algorithms are expected to be implemented in real-time, with each measurement coming in its own time. In fingerprinting, most existing works rely on fixed target localization and data sets consisting of RF features. However, in either the online or offline step, other sensor data like IMU and images can be used effectively with radio-based data sets. They can either narrow down the searching area in the online phase or directly be fused with fingerprints in the offline step. From a theoretical point of view, there is still room for localization improvement in accuracy and robustness in a data fusion scheme. For instance, to the author's best knowledge, there is no study on fusing promising measurements provided by 5G with images, LIDAR, etc. Several combinations of the sensor data and their performance under different situations and parameters need to be researched. %The 5G network parameters and BS configurations for different situations, such as indoor, outdoor, and NLOS conditions, are still unclear.

\subsection{Combination of Multiple Estimators}
Combining multiple estimators for range-based or  fingerprinting could compensate for each method's disadvantages, improve accuracy, and introduce resiliency to failure cases. This technique is well-established in the general machine learning task and is known as ensemble learning~\cite{zhou2012ensemble}. The ensemble method combines different estimators and often yields much more accurate results than individual methods into four main paradigms Bayesian averaging, error-correcting output coding, Bagging, and boosting~\cite{dietterich2000ensemble}. 

\subsection{Cooperative Localization}
 With V2X technology, many heterogeneous multi-robot applications comprising UGVs and UAVs can be envisioned. This enables more efficient use of robots for diverse scenarios like coverage, formation, task distribution, etc. For example, in a search and rescue operation, a group of heterogeneous robots may collaborate while some act as an anchor in the absence of a signal for some or all team members. On the other hand, sharing information by robots operating in the same area would endow robustness and accuracy to the localization system, as each robot can benefit from the knowledge and estimation of its neighbour(s).

\subsection{Orientation Estimation}
Deployment of a multi-array antenna system paves the way for orientation estimation. However, attitude estimation would be challenging, especially in 3D space while the UAV is flying. Therefore, integrating RF signals with other data to improve attitude estimation in 3D seems to be an attractive research topic left untouched. 

% Based on these phase pattern characteristics, researchers tend to determine the orientation of a user array antenna by searching over a set of orientation candidates for the desired value closest to the true orientation. The search result is determined by comparing the relative positioning solutions between the user array antenna and a reference station.

\subsection{Experimental setup and Realistic Simulation}
 Real experiments and setups for the use of 5G with robots are lacking in the state-of-the-art. This is critical as the actual setup's outcome does not necessarily match the predictions and expectations. Two examples are the offloading and handover in the 5G network. In the uplink, which is crucial for offloading, the data rate in 5G NR is supposed to be significantly improved, but \cite{albanese_first_2021} recorded the maximum of 67 Mbit/s in the uplink for a flying drone showing no improvement compared to 4G. The impact of handover is also worth further researching, especially for high mobility scenarios. \cite{albanese_first_2021} shows that the handover rate between LTE and 5G is too high, which is unacceptable.

\subsection{Off-loading} With offloading, more computation power will be available. This calls for new complex strategies and algorithms capable of being run and worked in parallel. For example, the onboard computer and edge server might work collaboratively and separately. While onboard computers process some part of the data to execute a rough localization, heavy algorithms are to be implemented on the edge server separately when there is no communication between the two processors. Both algorithms then need to be merged to output the refined localization result. At the same time, new strategies might be developed with the processing power on edge devices to focus more on accuracy and robustness instead of the computational burden. Most state-of-the-art localization tries to balance accuracy, complexity, and computation power. 

\subsection{Simultaneous Localization and Mapping} A critical task in robotics is Simultaneous Localization and Mapping (SLAM). As the robot moves in an unfamiliar environment, it needs to construct a map and localize itself on the map while navigating. The use of multi robots instead of a single one, while each shares its maps with others, is a fascinating topic that adds efficiency to the SLAM. Multi-robot SLAM (MRSLAM) is implemented in two ways. Using a central station to which all robots disseminate the collected data to do all the processing, construct the global map, and transmit this data back to each robot. In a decentralized system, each robot performs locally, and whenever they visit, other robots share their local map, based on which they update their map. 
Accordingly, one big challenge in centralized and distributed MRSLAM is the constraint on the communication bandwidth, limited computation power, and memory. 5G NR will circumvent these limitations by facilitating edge computing \cite{huang2022edge, huang2021colaslam}, providing accurate RF-based measurements, and enabling relative localization for each vehicle in the team via sidelink and V2V technologies.

\subsection{Vertical Localization Accuracy}
As discussed earlier, the localization accuracy in the z-direction is always less than in the x-y. Notably, the multi-path effect and the small offset of anchors in the z-direction yield poor vertical accuracy relying on conventional methods \cite{charlety20222d}. MIMO technology and wide-band mmWave system of 5G provide promising solutions to improve the 3D positioning, like taking full advantage of CFR in all frequencies or analysing antenna radiation pattern~\cite{sinha2022impact}.

\subsection{Safety} 
Safety is a major concern in robotics-related tasks. In general, there are two types of communication that apply to UAVs: Control and Non-Payload Communication (CNPC) and Payload Communication (PC). CNPC refers to control commands, way-points and navigation, usually in the order of several Kbps. Instead, the PC includes data transmission to the edge server, ground or aerial centre for processing. This information ranges from small-sized data, e.g., IMU, GPS, and RSS, to large-size high-resolution aerial images or LIDAR scans. While communication link reliability might not be vital for PC, avoiding communication interruptions is critical regarding CNPC. This leads to the concern of interruption of the CNPC link when there is no LOS path. Remarkably, using mmWave and beam-forming features of 5G call for additional care because they come with the disadvantage of very high propagation pass loss~\cite{hosseini2019uav}. Therefore, the design of trajectory and path-planning algorithms that guarantee LOS communication would be an interesting research topic. Another concern for CNPC links is cyber-physical attacks that might corrupt data transmission or cause the misaction of the UAVs or UGVs~\cite{mei_rss-based_2021}.

% time synchronization for slam fusing multiple sensors, fps and time to compute features, wide availability of development kit programmable for small robots. 

% The mobility and the issue of receiving diverse information at different frequencies and times is another challenge missing in the literature. Localization algorithms are expected to be implemented in real-time, with each measurement coming in its own time.